%% file: main.tex
\documentclass[10pt,journal,compsoc]{IEEEtran}
\usepackage{graphicx}

\ifCLASSOPTIONcompsoc
  \usepackage[nocompress]{cite}
\else
  \usepackage{cite}
\fi
\usepackage{hyperref}       % hyperlinks
\usepackage{url}            % simple URL typesetting
\usepackage{epsfig}
\usepackage{graphicx}
\usepackage{booktabs}       % professional-quality tables
\usepackage{amsfonts}       % blackboard math symbols
\usepackage{nicefrac}       % compact symbols for 1/2, etc.
\usepackage{microtype}      % microtypography
\usepackage[table]{xcolor}         % colors
\usepackage{amsmath}
\usepackage{amssymb}
\usepackage{cleveref}
\usepackage{subfig}
\usepackage{enumerate}
\usepackage{amsthm}
\usepackage{multirow}
\usepackage{wrapfig}
\usepackage{booktabs}
\usepackage{floatrow}
\usepackage{cancel}
\usepackage{makecell}
\usepackage[ruled,vlined]{algorithm2e}
\newtheorem{prop}{Proposition}

\newtheorem{thm}{Theorem}
\usepackage{enumitem}
\usepackage{listings}

\usepackage{xcolor}
\definecolor{codegreen}{rgb}{0,0.6,0}
\definecolor{codegray}{rgb}{0.5,0.5,0.5}
\definecolor{codepurple}{rgb}{0.58,0,0.82}
\definecolor{backcolour}{rgb}{0.95,0.95,0.92}

\lstdefinestyle{mystyle}{
    backgroundcolor=\color{backcolour},   
    commentstyle=\color{codegreen},
    keywordstyle=\color{magenta},
    numberstyle=\tiny\color{codegray},
    stringstyle=\color{codepurple},
    basicstyle=\ttfamily\footnotesize,
    breakatwhitespace=false,         
    breaklines=true,                 
    captionpos=b,                    
    keepspaces=true,                 
    numbers=left,                    
    numbersep=5pt,                  
    showspaces=false,                
    showstringspaces=false,
    showtabs=false,                  
    tabsize=2
}
\setlist[itemize]{leftmargin=*}
\newfloatcommand{capbtabbox}{table}[][\FBwidth]
\ifCLASSINFOpdf
\else
\fi

\usepackage{pifont}% http://ctan.org/pkg/pifont
\newcommand{\cmark}{\text{\ding{51}}}%
\newcommand{\xmark}{\text{\ding{55}}}%

\begin{document}
\title{RankFeat\&RankWeight: Rank-1 Feature/Weight Removal for Out-of-distribution Detection}
\author{Yue~Song,~\IEEEmembership{Member,~IEEE,}
        Wei~Wang,~\IEEEmembership{Member,~IEEE,}
        Nicu~Sebe,~\IEEEmembership{Senior Member,~IEEE}% <-this % stops a space
\IEEEcompsocitemizethanks{\IEEEcompsocthanksitem Yue Song and Nicu Sebe are with Department
of Information Engineering and Computer Science, University of Trento, Trento 38123,
Italy. Wei Wang is with Beijing Jiaotong University, Beijing, China. Wei Wang is the corresponding author.\\
E-mail: \{yue.song, nicu.sebe\}@unitn.it, wei.wang@bjtu.edu.cn}
%\IEEEcompsocthanksitem J. Doe and J. Doe are with Anonymous University.}% <-this % stops an unwanted space
\thanks{Manuscript received April 19, 2005; revised August 26, 2015.}}


\markboth{IEEE TRANSACTIONS ON PATTERN ANALYSIS AND MACHINE INTELLIGENCE}%
{Shell \MakeLowercase{\textit{et al.}}: Bare Demo of IEEEtran.cls for Computer Society Journals}



\IEEEtitleabstractindextext{%
\begin{abstract}
 The task of out-of-distribution (OOD) detection is crucial for deploying machine learning models in real-world settings. In this paper, we observe that the singular value distributions of the in-distribution (ID) and OOD features are quite different: the OOD feature matrix tends to have a larger dominant singular value than the ID feature, and the class predictions of OOD samples are largely determined by it. This observation motivates us to propose \texttt{RankFeat}, a simple yet effective \emph{post hoc} approach for OOD detection by removing the rank-1 matrix composed of the largest singular value and the associated singular vectors from the high-level feature. \texttt{RankFeat} achieves \emph{state-of-the-art} performance and reduces the average false positive rate (FPR95) by 17.90\% compared with the previous best method. The success of \texttt{RankFeat} motivates us to investigate whether a similar phenomenon would exist in the parameter matrices of neural networks. We thus propose \texttt{RankWeight} which removes the rank-1 weight from the parameter matrices of a single deep layer. Our \texttt{RankWeight}is also \emph{post hoc} and only requires computing the rank-1 matrix once. As a standalone approach, \texttt{RankWeight} has very competitive performance against other methods across various backbones. Moreover, \texttt{RankWeight} enjoys flexible compatibility with a wide range of OOD detection methods. The combination of \texttt{RankWeight} and \texttt{RankFeat} refreshes the new \emph{state-of-the-art} performance, achieving the FPR95 as low as 16.13\% on the ImageNet-1k benchmark. Extensive ablation studies and comprehensive theoretical analyses are presented to support the empirical results. Code is publicly available via \url{https://github.com/KingJamesSong/RankFeat}.
\end{abstract}

%Cover Letter and 
%In our previous conference paper, we propose two fast algorithms to compute the matrix square root. For the application of inverse square root, we solve the linear system to avoid the matrix inverse, which is more numerically stable and faster. However, solving the linear system adds extra computational burden to the training and might not be efficient than the In this extension, we target the drawback and extend our method to the case of inverse square root. 

}

\maketitle

\IEEEdisplaynontitleabstractindextext

% \fi
\IEEEpeerreviewmaketitle

\input{1_intro}
\input{2_related}
\input{3_method}
\input{4_exp}
\input{5_conclusion}
% or


%\ifCLASSOPTIONcompsoc
%\else
%\fi

\ifCLASSOPTIONcaptionsoff
  \newpage
\fi



%\bibliographystyle{IEEEtran}
\bibliographystyle{IEEEtran}
\bibliography{egbib}
%


\begin{IEEEbiography}[{\includegraphics[width=1in,height=1.25in,keepaspectratio]{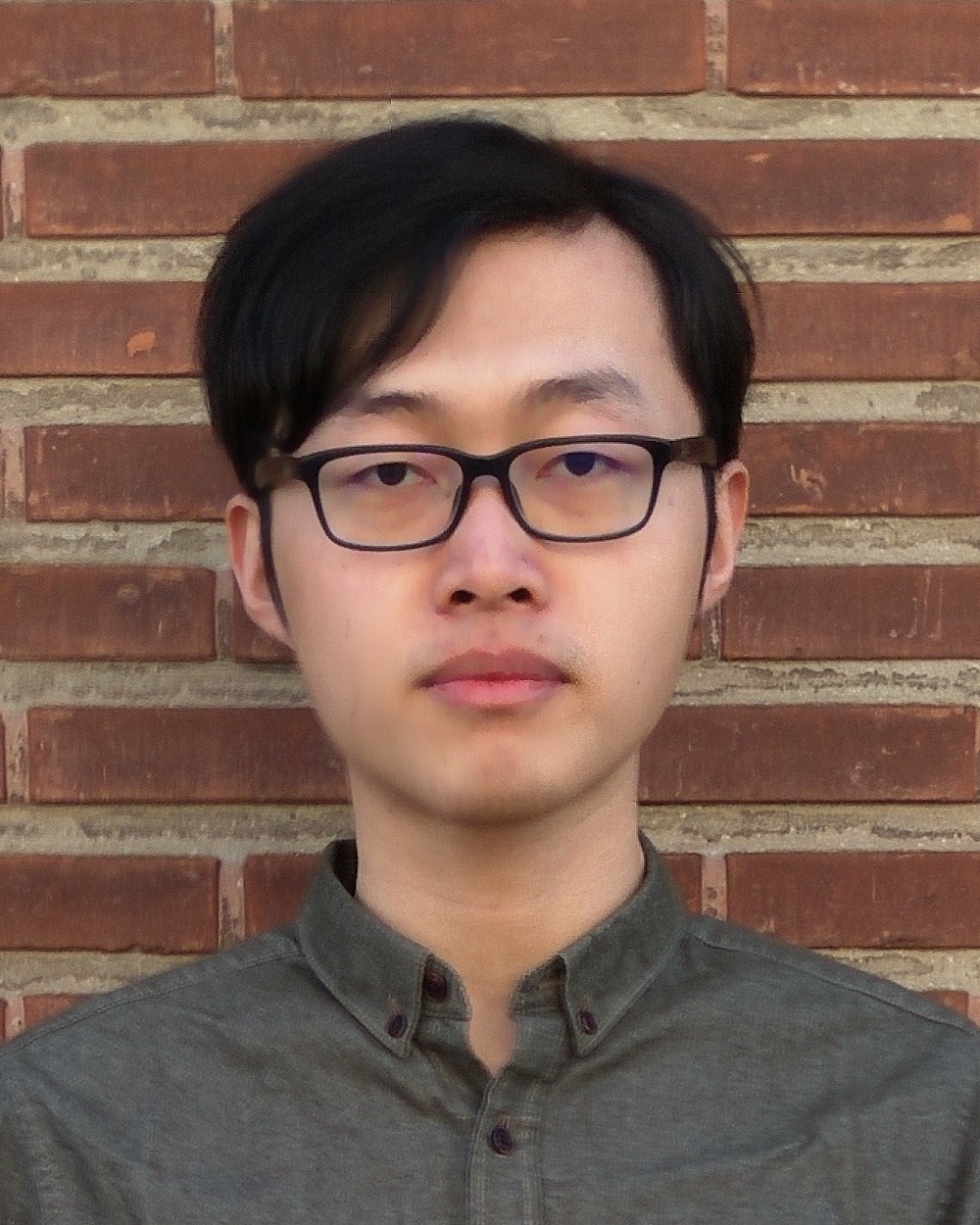}}]{Yue Song}
received the B.Sc. \emph{cum laude} from KU Leuven, Belgium and the joint M.Sc. \emph{summa cum laude} from the University of Trento, Italy and KTH Royal Institute of Technology, Sweden. Currently, he is a Ph.D. student with the Multimedia and Human Understanding Group (MHUG) at the University of Trento, Italy. His research interests are computer vision, deep learning, and numerical analysis and optimization.
\end{IEEEbiography}


\begin{IEEEbiography}[{\includegraphics[width=1in,height=1.25in,keepaspectratio]{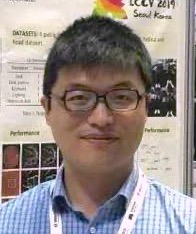}}]{Wei Wang}
is an Assistant Professor of Computer Science at University of Trento, Italy. Previously, after obtaining his PhD from University of
Trento in 2018, he became a Postdoc at EPFL,
Switzerland. His research interests include machine learning and its application to computer
vision and multimedia analysis.
\end{IEEEbiography}

\begin{IEEEbiography}[{\includegraphics[width=1in,height=1.25in,keepaspectratio]{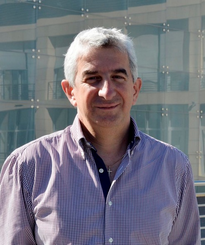}}]{Nicu Sebe} is Professor with the University of
Trento, Italy, leading the research in the areas
of multimedia information retrieval and human
behavior understanding. He was the General
Co- Chair of ACM Multimedia 2013, and the
Program Chair of ACM Multimedia 2007 and
2011, ECCV 2016, ICCV 2017 and ICPR 2020.
He is a fellow of the International Association for
Pattern Recognition.
\end{IEEEbiography}
%\newpage

\appendices
\newpage

\input{6_supp}

%\vfill


\end{document}

%% file: 1_intro.tex
\section{Introduction}

%OOD Detection 

In the real-world applications of deep learning, understanding whether a test sample belongs to the same distribution of training data is critical to the safe deployment of machine learning models. The main challenge stems from the fact that current deep learning models can easily give over-confident predictions for out-of-distribution (OOD) data~\cite{nguyen2015deep}. Recently a rich line of literature has emerged to address the challenge of OOD detection~\cite{wang2021can,huang2021importance,bibas2021single,diffenderfer2021winning,sun2021react,gibbs2021adaptive,fort2021exploring,sutter2021robust,ye2021towards,kumar2022fine,garg2022leveraging,zaeemzadeh2021out,du2022vos,gomes2022igeood,haroush2021statistical,wang2022vim,djurisic2023extremely,sun2022dice,zhang2023openood,park2023nearest,park2023understanding,guan2023revisit}.

\begin{figure*}[t]
    \centering
    \includegraphics[width=0.99\linewidth]{imgs/teaser2.pdf}
    \caption{\textbf{(a)} The distribution of top-5 singular values for the ID and OOD features on ImageNet-1k and SUN. The OOD feature matrix tends to have a significantly larger dominant singular value. \textbf{(b)} After removing the rank-1 matrix composed by the dominant singular value and singular vectors, the class predictions of OOD data are severely perturbed, while those of ID data are moderately influenced. This observation indicates that the decisions of OOD data heavily depend on the dominant singular value and the corresponding singular vectors of the feature matrix. In light of this finding, we get motivated to propose \texttt{RankFeat} for OOD detection by removing the rank-1 matrix  from the high-level feature. \textbf{(c)} After pruning the parameters of a single deep layer by removing the rank-1 matrix similarly, the class predictions of ID and OOD data exhibit distinct behaviors: most ID samples remain consistent class predictions, while OOD data is largely perturbed. This implies that the rank-1 parameter matrix of deep layers also plays a crucial role in making decisions about data samples. We thus propose \texttt{RankWeight} for \emph{post hoc} OOD detection by removing the rank-1 matrix from the deep parameter matrix of only one layer. The observations also hold for other OOD datasets.}
    \label{fig:cover}
\end{figure*}

%The singular value distribution of ID and OOD are quite different.
Previous OOD detection approaches either rely on the feature distance~\cite{lee2018simple}, activation abnormality~\cite{sun2021react}, or gradient norm~\cite{huang2021importance}. In this paper, we tackle the problem of OOD detection from another perspective: by analyzing the spectrum of the high-level feature matrices (\emph{e.g.,} the output of Block 3 or Block 4 of a typical ResNet~\cite{he2016deep} model), we observe that the feature matrices have quite different singular value distributions for the in-distribution (ID) and OOD data (see Fig.~\ref{fig:cover}(a)): \textit{the OOD feature tends to have a much larger dominant singular value than the ID feature, whereas the magnitudes of the remaining singular values are very similar.} This peculiar behavior motivates us to remove the rank-1 matrix composed of the dominant singular value and singular vectors from the feature. As displayed in Fig.~\ref{fig:cover}(b), removing the rank-1 feature drastically perturbs the class prediction of OOD samples; a majority of predictions have been changed. On the contrary, most ID samples have consistent classification results before and after removing the subspace. \textit{This phenomenon indicates that the over-confident prediction of OOD samples might be largely determined by the dominant singular value and the corresponding singular vectors.}

%OOD data tends to have larger dominant singular value and the over-confident prediction is largely caused by this singular value and the rank-1 matrix. Removing the rank-1 approximation of the feature can reveal the classification 
%Similar observation about the usefulness of rank-1 feature is also pointed out in~\cite{zaeemzadeh2021out}.

\begin{figure}[htbp]
    \centering
    \includegraphics[width=0.99\linewidth]{imgs/diagram.pdf}
    \caption{Visual illustration of our \texttt{RankFeat} and \texttt{RankWeight}: \texttt{RankFeat} removes the rank-1 \textbf{feature matrix} from the deep layers, while \texttt{RankWeight} perturbs the \textbf{weight matrix} of deep layers by removing the rank-1 subspace similarly.}
    \label{fig:diagram}
\end{figure}

%In light of this finding, 
Based on this observation, we \textbf{assume} that the first singular value of the OOD feature tends to be much larger than that of the ID feature. The intuition behind this is that the OOD feature corresponds to a larger Principal Component Analysis (PCA) explained variance ratio (being less informative), and the well-trained network weights might cause and amplify the difference (see Sec. D of the supplementary for the detailed illustration). Hence, we conjecture that leveraging this gap might help to better distinguish ID and OOD samples. To this end, we propose \texttt{RankFeat}, a simple but effective \textit{post hoc} approach for OOD detection. \texttt{RankFeat} perturbs the high-level feature by removing its rank-1 matrix composed of the dominant singular value and vectors. Then the logits derived from the perturbed features are used to compute the OOD score function. By removing the rank-1 feature, the over-confidence of OOD samples is mitigated, and consequently the ID and OOD data can be better distinguished (see Fig.~\ref{fig:score_dist}). Our \texttt{RankFeat} establishes the \textit{state-of-the-art} performance on the large-scale ImageNet benchmark and a suite of widely used OOD datasets across different network depths and architectures. In particular, \texttt{RankFeat} outperforms the previous best method by \textbf{17.90\%} in the average false positive rate (FPR95) and by \textbf{5.44\%} in the area under curve (AUROC). Extensive ablation studies are performed to reveal important insights of \texttt{RankFeat}, and comprehensive theoretical analyses are conducted to explain the working mechanism.

The success of \texttt{RankFeat} implies that a similar abnormality of the rank-1 matrix might also exist in other aspects of deep learning models such as the parameter matrices of deep network layers. To investigate this research question, we also measure the impact of the rank-1 parameter matrix on the model decisions. As can be seen from Fig.~\ref{fig:cover}(c), the class predictions are also largely dependent on the rank-1 matrix of deep parameters; pruning the parameter matrix would heavily perturb the model decisions. This observation implies that the over-confidence of OOD samples might be closely related to the rank-1 parameter matrix. In light of this finding, we propose \texttt{RankWeight}, another simple yet effective \emph{post hoc} OOD detection approach which removes the rank-1 weight matrix from the last deep parameter matrix before the fully-connected layer. Fig.~\ref{fig:diagram} illustrates the working mechanisms of our \texttt{RankFeat} and \texttt{RankWeight}. Compared with \texttt{RankFeat}, \texttt{RankWeight} is more computationally efficient as it only requires (approximate) matrix decomposition once for only one layer of the model. When \texttt{RankWeight} is used standalone, it establishes competitive performance against previous baselines. Moreover, our \texttt{RankWeight} can be integrated with a wide range of OOD detection methods and boost their performance, including \texttt{GradNorm}~\cite{huang2021importance}, \texttt{ReAct}~\cite{sun2021react}, \texttt{ASH}~\cite{djurisic2023extremely}, \texttt{VRA}~\cite{xu2023vra}, and our \texttt{RankFeat}. For example, combing \texttt{RankWeight} and \texttt{RankFeat} attains an FPR95 of \textbf{16.13\%} and an AUROC of \textbf{96.20\%}, and integrating \texttt{RankWeight} into \texttt{ASH}~\cite{djurisic2023extremely} attains an FPR95 of \textbf{15.46\%} and an AUROC of \textbf{97.13\%}, both of which achieve new \emph{state-of-the-art} performance across datasets on ImageNet-1k benchmark. We conduct some ablation studies to understand the impact of pruning different layers using \texttt{RankWeight}. Also, some theoretical analyses are performed to show that performing \texttt{RankWeight} can tighten the score upper bound.

%we check the eigenspectrum of the deep parameter matrices. 

%Despite the \emph{post hoc} flexibility, our \texttt{RankFeat} suffers from the limitation of  and cannot adapt the model when the data is available. To resolve this issue, we introduce two variants of our method: \texttt{RankFeat+} and \texttt{RankFeat++} that involve ID and OOD sets as auxiliary regularization for light-weighted fine-tuning, respectively. For \texttt{RankFeat+}, we apply rank-1 feature removal on the ID set and push the corresponding logits to stay away from the uniform distribution, therefore exposing the inlier to the model. For \texttt{RankFeat++}, we also apply rank-1 feature removal on the OOD set but push the corresponding logits to stay close to the uniform distribution, thus performing outlier exposure and making the model aware of the OOD samples. Within only $1$ epoch of fine-tuning on ID/OOD datasets, our method can further improve the performance on the large-scale OOD benchmark by a large margin. Specifically,  

\begin{figure*}[t]
    \centering
    \includegraphics[width=0.99\linewidth]{imgs/score_dist_3rows.pdf}
    \caption{The score distributions of \texttt{Energy}~\cite{liu2020energy} (top row), our \texttt{RankFeat} (middle row), and our proposed \texttt{RankFeat}+\texttt{RankWeight} (bottom row) on four OOD datasets. Our method can better separate the ID and OOD data. }
    \label{fig:score_dist}
\end{figure*}

The \textbf{key results and main contributions} are threefold:
\begin{itemize}
    \item We propose \texttt{RankFeat} and \texttt{RankWeight}, two simple yet effective \textit{post hoc} approaches for OOD detection by removing the rank-1 matrix from the high-level feature and deep parameter matrices, respectively. \texttt{RankFeat} achieves the \textit{state-of-the-art} performance across benchmarks and models, reducing the average FPR95 by \textbf{17.90\%} and improving the average AUROC by \textbf{5.44\%} compared to the previous best method. Combining \texttt{RankFeat} and \texttt{RankWeight} further establishes \emph{state-of-the-art} performance on the ImageNet-1k benchmark, achieving a low FPR95 of $\textbf{16.13\%}$ and a high AUROC of $\textbf{96.20\%}$.
    \item We perform extensive ablation studies to illustrate the impact of (1) removing or keeping the rank-1 feature, (2) removing the rank-n feature (n$>$1), (3) applying our \texttt{RankFeat} at various network depths, (4) the number of iterations to iteratively derive the approximate rank-1 matrix for acceleration but without performance degradation, (5) different fusion strategies to combine multi-scale features for further performance improvements, and (6) pruning different number of layers using \texttt{RankWeight}.
    %Moreover, we show that (4) our method can be accelerated by the iterative technique for acceleration without performance degradation, and (5) the multi-scale features can be fused for  
    \item Comprehensive theoretical analyses are conducted to explain the working mechanism and to underpin the superior empirical results. We show that (1) removing the rank-1 feature reduces the upper bound of OOD scores more, (2) removing the rank-1 matrix makes the statistics of OOD feature closer to random matrices, (3) both \texttt{RankFeat} and \texttt{ReAct}~\cite{sun2021react} work by optimizing the upper bound containing the largest singular value. \texttt{ReAct}~\cite{sun2021react} indirectly and manually clips the underlying term, while \texttt{RankFeat} directly subtracts it, and (4) our \texttt{RankWeight} can effectively tighten the score upper bound.
\end{itemize}

This paper is an extension of the previous conference paper~\cite{song2022rankfeat}. In~\cite{song2022rankfeat}, we observe the 
abnormality of the rank-1 feature and propose \texttt{RankFeat} for OOD detection by removing the rank-1 matrix from the high-level feature maps. The journal extension validates that similar behavior can be also observed in the parameters of deep network layers, and further proposes \texttt{RankWeight} which performs OOD detection by pruning the rank-1 weight from the deep parameter matrices. Our \texttt{RankWeight} can be combined with many OOD detection methods, and particularly jointly applying \texttt{RankFeat} and \texttt{RankWeight} establishes the new \emph{state-of-the-art} on the large-scale OOD detection benchmark. We also conduct some theoretical analysis and ablation studies to improve the understanding of \texttt{RankWeight}.

The rest of the paper is organized as follows: Sec.~\ref{sec:related} describes the related work in distribution shifts and OOD detection. Sec.~\ref{sec:method} introduces our proposed \texttt{RankFeat} and \texttt{RankWeight} which remove the rank-1 matrix from the deep feature maps and parameter matrices respectively, and Sec.~\ref{sec:theory} presents the theoretical analyses that underpin our methods.  Sec.~\ref{sec:exp} provides experimental results and some in-depth analysis. Finally, Sec.~\ref{sec:conc} summarizes the conclusions.

%% file: 2_related.tex
\section{Related Work}
\label{sec:related}
\subsection{Distribution Shifts} 

Distribution shifts have been a long-standing problem in the machine learning research community~\cite{hand2006classifier,quinonero2008dataset,koh2021wilds,wiles2021fine}. The problem of distribution shifts can be generally categorized as shifts in the input space and shifts in the label space. Shifts only in the input space are often deemed as \emph{covariate shifts}~\cite{hendrycks2018benchmarking,ovadia2019can}. In this setting, the inputs are corrupted by perturbations or shifted by domains, but the label space stays the same~\cite{hsu2020generalized,sun2020test}. The aim is mainly to improve the robustness and generalization of a model~\cite{hendrycks2019augmix}. For OOD detection, the labels are disjoint and the main concern is to determine whether a test sample should be predicted by the pre-trained model~\cite{liang2017enhancing,hsu2020generalized}. 

Some related sub-fields also tackle the problem of distribution shifts in the label space, such as novel class discovery~\cite{han2019learning,zhong2021neighborhood}, open-set recognition~\cite{scheirer2012toward,vaze2021open}, and novelty detection~\cite{abati2019latent,tack2020csi}. These sub-fields target specific distribution shifts (\emph{e.g.,} semantic novelty), while OOD encompasses all forms of shifts.

%They are closely related to the OOD detection but there are some subtle differences.
%\noindent \textbf{Related Sub-fields.}

\subsection{OOD Detection with Discriminative Models} 

The early work on discriminative OOD detection dates back to the classification model with rejection option~\cite{chow1970optimum,fumera2002support}. The OOD detection methods can be generally divided into training-need methods and \emph{post hoc} approaches. Training-needed methods usually need auxiliary OOD data to regularize the model for lower confidence or higher energy~\cite{lee2018training,malinin2018predictive,geifman2019selectivenet,hein2019relu,meinke2020towards,jeong2020ood,van2020uncertainty,yang2021semantically,wei2022mitigating,du2022unknown,katz2022training}. Compared with training-needed approaches, \emph{post hoc} methods do not require any extra training processes and could be directly applied to any pre-trained models. For the wide body of research on OOD detection, please refer to~\cite{yang2021generalized} for the comprehensive survey. Here we highlight some representative \emph{post hoc} methods. Nguyen~\emph{et al.}~\cite{nguyen2015deep} first observed the phenomenon that neural networks easily give over-confident predictions for OOD samples. The following research attempted to improve the OOD uncertainty estimation by proposing ODIN score~\cite{liang2017enhancing}, OpenMax score~\cite{bendale2016towards}, Mahalanobis distance~\cite{lee2018simple}, and Energy score~\cite{liu2020energy}.
Huang~\emph{et al.}~\cite{huang2021mos} pointed out that the traditional CIFAR benchmark does not extrapolate to real-world settings and proposed a large-scale ImageNet benchmark. More recently, Sun~\emph{et al.}~\cite{sun2021react} and Huang~\emph{et al.}~\cite{huang2021importance} proposed to tackle the challenge of OOD detection from the lens of activation abnormality and gradient norm, respectively. Following~\cite{sun2021react}, recent works emerge to find the optimal activation shaping methods from various perspectives~\cite{sun2022dice,djurisic2023extremely,xu2023vra}. In contrast, based on the empirical observation about the impact of the dominant singular pairs on model decisions, we propose \texttt{RankFeat} and \texttt{RankWeight}, two simple yet effective \emph{post hoc} solutions by removing the rank-1 subspace from the high-level features and deep parameter matrices, respectively. The technique of learning matrices has also been widely explored in various tasks of deep learning~\cite{fang2018flexible,fang2020dynamic,han2020projective}.
%The most similar work with ours is probably~\cite{zaeemzadeh2021out}. They forces the ID samples to be embedded into a union of first eigenvector subspace during training and computes the minimum angular distance as the score, which needs extra training process and the whole ID set. Therefore, their work does not belong to the \emph{post hoc} category.
%In contrast, our work is based on the empirical observation of singular value distributions of high-level features. From the perspective of matrix analysis, we proposes a simple yet effective solution by removing the rank-1 feature subspace from the representation. 

\subsection{OOD Detection with Generative Models} 

Different from discriminative models, generative models detect the OOD samples by estimating the probability density function~\cite{kingma2013auto,tabak2013family,rezende2014stochastic,van2016conditional,dinh2016density,huang2017stacked,bibas2021single,jiang2021revisiting}. A sample with a low likelihood is deemed as OOD data. Recently, a multitude of methods have utilized generative models for OOD detection~\cite{ren2019likelihood,serra2019input,wang2020further,xiao2020likelihood,kirichenko2020normalizing,schirrmeister2020understanding,kim2021locally}. Recent approaches focus on leveraging the power of generative models for synthesizing virtual outliers to regularize the decision boundary of ID and OOD samples~\cite{du2022vos,tao2023non,du2023dream}. However, as pointed out in~\cite{nalisnick2018deep}, generative models could assign a high likelihood to OOD data. Furthermore, generative models can be prohibitively harder to train and optimize than their discriminative counterparts, and the performance is often inferior.

%This might limit their practical usage. More recently, large vision language models

%% file: 3_method.tex
\section{RankFeat\&RankWeight: Rank-1 Feature/Weight Removal for OOD Detection}
\label{sec:method}

In this section, we introduce the background of OOD detection task and our proposed \texttt{RankFeat} and \texttt{RankWeight} that perform the OOD detection by removing the rank-1 matrix from the high-level feature and the deep parameter matrix, respectively.

\subsection{Preliminary: OOD Detection} 

The OOD detection is often formulated as a binary classification problem with the goal to distinguish between ID and OOD data. Let $f$ denote a model trained on samples from the ID data $\mathcal{D}_{in}$. For the unseen OOD data $\mathcal{D}_{out}$ at test time, OOD detection aims to define a decision function $\mathcal{G}(\cdot)$:
\begin{equation}
    \mathcal{G}(\mathbf{x})=
    \begin{cases}
    {\rm in} & \mathcal{S}(\mathbf{x})>\gamma ,\\
    {\rm out} & \mathcal{S}(\mathbf{x})<\gamma .
    \end{cases}
\end{equation}
where $\mathbf{x}$ denotes the data encountered at the inference stage, $\mathcal{S}(\cdot)$ is the seeking scoring function, and $\gamma$ is a chosen threshold to make a large portion of ID data correctly classified (\emph{e.g.,} $95\%$). The difficulty of OOD detection lies in designing an appropriate scoring function $\mathcal{S}(\cdot)$ such that the score distributions of ID and OOD data overlap as little as possible. 

\subsection{RankFeat: Rank-1 Feature Removal} 

Consider the reshaped high-level feature map $\mathbf{X}{\in}\mathbb{R}^{C{\times}HW}$ of a deep network (the batch size is omitted for simplicity). Here 'high-level feature' denotes the feature map that carries rich semantics in the later layers of a network (\emph{e.g.,} the output of Block 3 or Block 4 of a typical deep model like ResNet). 
Our \texttt{RankFeat} first performs the Singular Value Decomposition (SVD) on each individual feature matrix in the mini-batch to decompose the feature:
\begin{equation}
    \mathbf{X} = \mathbf{U}\mathbf{S}\mathbf{V}^{T}
    \label{eq:X}
\end{equation}
where $\mathbf{S}{\in}\mathbb{R}^{C{\times}HW}$ is the rectangular diagonal singular value matrix, and $\mathbf{U}{\in}\mathbb{R}^{C{\times}C}$ and $\mathbf{V}{\in}\mathbb{R}^{HW{\times}HW}$ are left and right orthogonal singular vector matrices, respectively. Then \texttt{RankFeat} removes the rank-1 matrix from the feature as:
\begin{equation}
    \mathbf{X}' = \mathbf{X} - \mathbf{s}_{1}\mathbf{u}_{1}\mathbf{v}_{1}^{T}
    \label{eq:feature_removal}
\end{equation}
where $\mathbf{s}_{1}$ is the largest singular value, and $\mathbf{u}_{1}$ and $\mathbf{v}_{1}$ are the corresponding left and right singular vectors, respectively. The perturbed feature is fed into the rest of the network to generate the logit predictions $\mathbf{y}'$. Finally, \texttt{RankFeat} computes the energy score of the logits for the input $\mathbf{x}$ as:%The matrix $\mathbf{s}_{1}\mathbf{u}_{1}\mathbf{v}_{1}^{T}$ presents the best rank-1 approximation of the feature.
\begin{equation}
    {\texttt{RankFeat}}(\mathbf{x})=\log\sum\exp(\mathbf{y}')
\end{equation}
By removing the rank-1 matrix composed by the dominant singular value $\mathbf{s}_{1}$, the over-confident predictions of OOD data are largely perturbed. In contrast, the decisions of ID data are mildly influenced. This could help to separate the ID and OOD data better in the logit space.

\begin{figure}[htbp]
    \centering
    \includegraphics[width=0.99\linewidth]{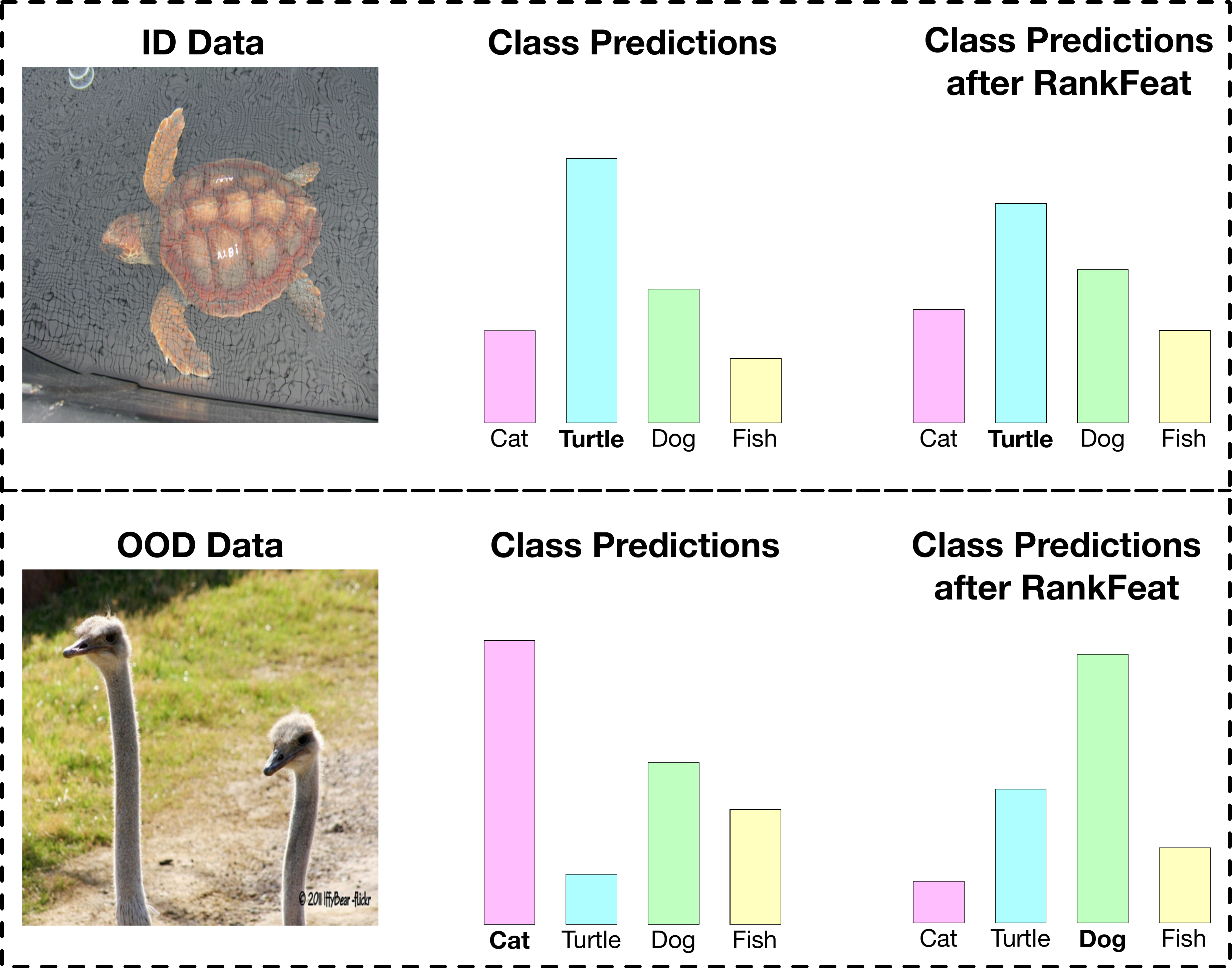}
    \caption{Impact of \texttt{RankFeat} on the class predictions of ID and OOD data. The class predictions of OOD data are significantly more perturbed than those of ID data.}
    \label{fig:rankfeat_example}
\end{figure}

Fig.~\ref{fig:rankfeat_example} gives an intuitive example of how the class predictions are impacted by our \texttt{RankFeat}. For the ID data, the class predictions are only mildly influenced; the same category is predicted correctly. By contrast, the logit distribution of OOD data is significantly perturbed, and the predicted category is inconsistent after performing \texttt{RankFeat}.

%Since the SVD is known as the best low-rank approximation technique~\cite{eckart1936approximation}, the matrix $\mathbf{s}_{1}\mathbf{u}_{1}\mathbf{v}_{1}^{T}$ presents the best rank-1 approximation of the feature.  

%Typically the channel number $C$ is much larger than the product spatial dimension $HW$ of the feature map. 

%\noindent \textbf{RankFeat: Removing Rank-1 Feature Matrix.} 
%\noindent \textbf{OOD detection with RankFeat.}

\subsubsection{Acceleration by Power Iteration} 

Since \texttt{RankFeat} only involves the dominant singular value and vectors, there is no need to compute the full SVD of the feature matrix. Hence our method can be potentially accelerated by Power Iteration (PI). The PI algorithm is originally used to approximate the dominant eigenvector of a Hermitian matrix. With a slight modification, it can also be applied to general rectangular matrices. Given the feature $\mathbf{X}$, the modified PI takes the coupled iterative update:
\begin{equation}
    \mathbf{v}_{k}=\frac{\mathbf{X}\mathbf{u}_{k}}{||\mathbf{X}\mathbf{u}_{k}||},\  \mathbf{u}_{k+1}=\left(\frac{\mathbf{v}_{k}^{T}\mathbf{X}}{||\mathbf{v}_{k}^{T}\mathbf{X}||}\right)^{T}
\end{equation}
where $\mathbf{u}_{0}$ and $\mathbf{v}_{0}$ are initialized with random orthogonal vectors and converge
to the left and right singular vectors, respectively. After certain iterations, the dominant singular value is computed as $\mathbf{s}_{1}{=}\mathbf{v}_{k}^{T}\mathbf{X}\mathbf{u}_{k}$. As will be illustrated in Sec.~\ref{sec:pi_com}, the approximate solution computed by PI achieves very competitive performance against the SVD but yields much less time overhead.

\subsubsection{Combination of Multi-scale Features} 

Our \texttt{RankFeat} works at various later depths of a model, \emph{i.e.,} Block 3 and Block 4. Since intermediate features might focus on different semantic information, their decision cues are very likely to be different. It is thus natural to consider fusing the scores to leverage the distinguishable information of both features for further performance improvements. Let $\mathbf{y}'$ and $\mathbf{y}''$ denote the logit predictions of Block 3 and Block 4 features, respectively. \texttt{RankFeat} performs the fusion at the logit space and computes the score function as $\log\sum\exp(\nicefrac{(\mathbf{y}'+\mathbf{y}'')}{2})$. Different fusion strategies will be explored in Sec.~\ref{sec:fusion}.

\subsection{RankWeight: Rank-1 Weight Removal} 

The impressive performance of \texttt{RankFeat} indicates that the over-confidence of OOD samples is related to the rank-1 feature map composed by the dominant singular pairs. A natural question is \textit{can we observe similar behaviors in other aspects of deep models and further leverage the abnormality for OOD detection?} We investigate this research question and find that the rank-1 parameter matrix also has a significant impact on the decisions of OOD data. We propose our \texttt{RankWeight} which similarly removes the rank-1 weight from the parameter matrices of the deep layer.

Let $\mathbf{M}$ denote the parameter matrix at the deep layer which generates the high-level feature map $\mathbf{X}$ in Eq.~(\ref{eq:X}). In convolution neural networks $\mathbf{M}$ refers to the weights of convolution kernels, while $\mathbf{M}$ represents the weights of fully-connected layers in Transformers. Similar to Eqs.~(\ref{eq:X}) and~(\ref{eq:feature_removal}), we first perform SVD on the weights and remove the rank-1 matrix from the parameter as:
\begin{equation}
    \mathbf{M}' = \mathbf{M} - \Tilde{\mathbf{s}}_{1}\Tilde{\mathbf{u}}_{1}\Tilde{\mathbf{v}}_{1}^{T}
\end{equation}
where $\Tilde{\mathbf{s}}_{1}$, $\Tilde{\mathbf{u}}_{1}$, and $\Tilde{\mathbf{v}}_{1}$ are the dominant singular value and vectors of $\mathbf{M}$. For a given input $\mathbf{x}$, the model with perturbed parameters $\mathbf{M}$ predicts the logits $\Tilde{\mathbf{y}}$. We compute the score function as:
\begin{equation}
    {\texttt{RankWeight}}(\mathbf{x})=\log\sum\exp(\Tilde{\mathbf{y}})
\end{equation}
Our method prunes the weights that are likely to cause the over-confidence of OOD samples. As can be seen in Fig.~\ref{fig:cover}(c), the pruning procedure has a slight impact on ID data but drastically perturbs the predictions of OOD samples. The ID and OOD data are thus easier to separate in the logit space.

\subsubsection{Layers to Prune}

We perform \texttt{RankWeight} \textbf{at the last parametric layer before the fully-connected layer} of the model, {\emph{i.e.,} the last convolution layer for convolution networks or the linear layer of the last MLP for Transformers. We find that it is sufficient to prune the parameters of \textbf{only one layer} in the model using \texttt{RankWeight}. As will be illustrated in Sec.~\ref{sec:pruning}, pruning more layers would negatively influence the performance. This observation might also indicate that the over-confidence of OOD samples happens more in the deep layers.

\subsubsection{No Incurred Time Costs}

Our \texttt{RankWeight} only requires computing the SVD of the parameter matrices once during the test stage. The time cost is negligible compared to classifying the whole ID dataset and multiple OOD test sets. Therefore, it almost does not incur any additional time costs to the standard inference procedure. In practice, the computational cost for each sample is the same as a single forward pass.

%Notice that our approach is very time-efficient. 

%It does not incur additional time costs than the normal inference. 

\subsubsection{Combination with RankFeat}

Our \texttt{RankWeight} can be further combined with \texttt{RankFeat} to perform joint OOD detection. That is, we simultaneously remove the rank-1 matrix from both the high-level feature and the deep parameter matrix. These two methods complement each other and eliminate the over-confidence originating from the feature and the parameters.

\subsubsection{Compatibility with Other OOD Approaches}

Notably, our \texttt{RankWeight} can be incorporated with a wide range of OOD detection methods, including GradNorm~\cite{huang2021importance}, ReAct~\cite{sun2021react}, ASH~\cite{djurisic2023extremely}, and VRA~\cite{xu2023vra}. As will be discussed in Sec.~\ref{sec:rankweight_integration}, the integration of \texttt{RankWeight} would further boost the performance of these methods by a large margin of \textbf{18.83\%} in FPR95 on average.   

Notice that our \texttt{RankWeight} simply perturbs the model weights of one layer, which is not data-aware. The standalone use of \texttt{RankWeight} has competitive performance on different benchmarks. Only when integrated into data-aware OOD detection methods as a plugin, it can boost the performance of other methods to new \textit{state-of-the-art}.
%and improve their performance. 

%\subsection{RankFeat++: OOD Rank-1 Feature Exposure}

%\subsection{RankFeat+++: ID and OOD Rank-1 Feature Exposure}

%% file: 4_exp.tex
\section{Theoretical Analysis}
\label{sec:theory}

In this section, we perform some theoretical analyses on \texttt{RankFeat} and \texttt{RankWeight} to support the empirical results. We start by proving that removing the rank-1 feature with a larger $\mathbf{s}_{1}$ would reduce the upper bound of \texttt{RankFeat} score more. Then based on Random Matrix Theory (RMT), we show that removing the rank-1 matrix makes the statistics of OOD features closer to random matrices. Subsequently, the theoretical connection of \texttt{ReAct} and our \texttt{RankFeat} is analyzed and discussed: both approaches work by optimizing the score upper bound determined by $\mathbf{s}_{1}$. \texttt{ReAct} manually uses a pre-defined threshold to clip the term with $\mathbf{s}_{1}$, whereas our \texttt{RankFeat} directly optimizes the bound by subtracting this term. Finally, we show our \texttt{RankWeight} can tighten the upper bound and make our bound analysis more practical.  

%Then the justification of why the largest singular value embeds meaningful information based on Random Matrix Theory (RMT) is presented.

\subsection{RankFeat Reduces Upper Bounds of OOD Data More} 

For our \texttt{RankFeat} score function, we can express its upper bound in an analytical form. Moreover, the upper bound analysis explicitly indicates that removing the rank-1 matrix with a larger first singular value would reduce the upper bound more. Specifically, we have the following proposition.

\begin{prop}
\label{prop:up_bound}
The upper bound of \texttt{RankFeat} score is defined as $\texttt{RankFeat}(\mathbf{x}) <\frac{1}{HW} \Big(\sum_{i=1}^{N} \mathbf{s}_{i} - \mathbf{s}_{1}\Big) ||\mathbf{W}||_{\infty} +  ||\mathbf{b}||_{\infty} + \log(Q)$ where $Q$ denotes the number of categories, and $\mathbf{W}$ and $\mathbf{b}$ are the weight and bias of the last layer, respectively. A larger $\mathbf{s}_{1}$ would reduce the upper bound more.
\end{prop}

\begin{proof}

For the feature $\mathbf{X}{\in}\mathbb{R}^{C{\times}HW}$, its SVD  $\mathbf{U}\mathbf{S}\mathbf{V}^{T}{=}\mathbf{X}$ can be expressed as the summation of rank-1 matrices $\mathbf{X}{=}\sum\mathbf{s}_{i}\mathbf{u}_{i}\mathbf{v}_{i}^{T}$. The feature perturbed by \texttt{RankFeat} can be computed as:
\begin{equation}
    \mathbf{X}' = \mathbf{X} - \mathbf{s}_{1}\mathbf{u}_{1}\mathbf{v}_{1}^{T} = \sum_{i=2}^{N} \mathbf{s}_{i}\mathbf{u}_{i}\mathbf{v}_{i}^{T}
    \label{eq:bound_x}
\end{equation}
where $N$ denotes the shorter side of the matrix (usually $N{=}HW$). In most deep models~\cite{he2016deep,he2016identity}, usually the last feature map needs to pass a Global Average Pooling (GAP) layer to collapse the width and height dimensions. The GAP layer can be represented by a vector
\begin{equation}
    \mathbf{m}=\frac{1}{HW}\begin{bmatrix}1,1,\dots,1\end{bmatrix}^{T}
\end{equation}
The pooled feature map is calculated as $\mathbf{X}'\mathbf{m}$. Then the output logits are computed by the matrix-vector product with the classification head as:
\begin{equation}
\begin{gathered}
    \mathbf{y}'=\mathbf{W}\mathbf{X'}\mathbf{m}+\mathbf{b} = \sum_{i=2}^{N} (s_{i} \mathbf{W}\mathbf{u}_{i}\mathbf{v}_{i}^{T}\mathbf{m}) + \mathbf{b} 
    \label{eq:logit}
\end{gathered}
\end{equation}
where $\mathbf{W}{\in}\mathbb{R}^{W{\times}C}$ denotes the weight matrix, $\mathbf{b}{\in}\mathbb{R}^{Q{\times}1}$ represents the bias vector, and $\mathbf{y}'{\in}\mathbb{R}^{Q{\times}1}$ is the output logits that correspond to the perturbed feature $\mathbf{X}'$. Our \texttt{RankFeat} score is computed as:
\begin{equation}
    \texttt{RankFeat}(\mathbf{x}) = \log \sum_{i=1}^{Q} \exp(\mathbf{y}'_{i})
    \label{eq:rankfeat}
\end{equation}
where $\mathbf{x}$ is the input image, and $Q$ denotes the number of categories. Here we choose Energy~\cite{liu2020energy} as the base function due to its theoretical alignment with the input probability density and its strong empirical performance. Eq.~\eqref{eq:rankfeat} can be re-formulated by the \texttt{Log-Sum-Exp} trick
\begin{equation}
\begin{gathered}
     \log \sum_{i=1}^{Q} \exp(\mathbf{y}'_{i}) =  \log \sum_{i=1}^{Q} \exp(\mathbf{y}'_{i}-\max(\mathbf{y}')) + \max(\mathbf{y}')\\
     %\max(\mathbf{y})   < \log \sum \exp(\mathbf{y}) < \max(\mathbf{y}) + \log(C)
\end{gathered}
\end{equation}
The above equation directly yields the tight bound as:
\begin{equation}
    \max(\mathbf{y}')   < \log \sum \exp(\mathbf{y}') < \max(\mathbf{y}') + \log(Q)
\end{equation}
%Suppose $\mathbf{y}'$ has at least one positive element.\footnote{This assumption holds in practice because the maximum of logits $y$ is (almost) always larger than $0$. According to our observation, there does not exist any all-negative vector $\mathbf{y}$ among all the ID and OOD datasets.}  
Since $\max(\mathbf{y}'){\leq}\max(|\mathbf{y}'|){=}||\mathbf{y}'||_{\infty}$, we have
\begin{equation}
\begin{aligned}
\texttt{RankFeat}(\mathbf{x}) &= \log \sum \exp(\mathbf{y}') \\
&< \max(\mathbf{y}') + \log(Q)\\
&\leq  ||\mathbf{y}'||_{\infty} + \log(Q)
    \label{eq:score_upper}
\end{aligned}
\end{equation}
The vector norm has the property of triangular inequality, \emph{i.e.,} $||\mathbf{a}+\mathbf{c}||\leq||\mathbf{a}||+||\mathbf{c}||$ holds for any vectors $\mathbf{a}$ and $\mathbf{c}$. Moreover, since both $\mathbf{u}$ and $\mathbf{v}$ are orthogonal vectors, we have the relation $||\mathbf{u}_{i}||_{\infty}{\leq}1$ and $||\mathbf{v}_{i}||_{\infty}{\leq}1$.
Relying on these two properties, injecting eq.~\eqref{eq:logit} into eq.~\eqref{eq:score_upper} leads to
\begin{equation}
\begin{aligned}
\texttt{RankFeat}(\mathbf{x}) &< \sum_{i=2}^{N} \mathbf{s}_{i} ||\mathbf{W}\mathbf{u}_{i}\mathbf{v}_{i}^{T}\mathbf{m}||_{\infty} {+} ||\mathbf{b}||_{\infty} {+} \log(Q)\\
&\leq \sum_{i=2}^{N} \mathbf{s}_{i} ||\mathbf{W}\mathbf{m}||_{\infty} {+}  ||\mathbf{b}||_{\infty} {+} \log(Q)
\end{aligned}
\end{equation}
 %Then the above inequality can be simplified as:
%\begin{equation}
%\texttt{RankFeat}(\mathbf{X}') < \sum_{i=2}^{N} \mathbf{s}_{i} ||\mathbf{W}\mathbf{m}||_{\infty} +  ||\mathbf{b}||_{\infty} + \log(Q) 
%\end{equation}
Since $\mathbf{m}$ is a scaled all-ones vector, we have $||\mathbf{W}\mathbf{m}||_{\infty}{=}\nicefrac{||\mathbf{W}||_{\infty}}{HW}$. The bound is simplified as:
\begin{equation}
    \texttt{RankFeat}(\mathbf{x}){<}\frac{1}{HW} \left(\sum_{i=1}^{N} \mathbf{s}_{i} - \mathbf{s}_{1}\right) ||\mathbf{W}||_{\infty} +  ||\mathbf{b}||_{\infty} + \log(Q)
\end{equation}
As indicated above, removing a larger $\mathbf{s}_{1}$ would reduce the upper bound of \texttt{RankFeat} score more.
%Now the data-dependent variables only involve the singular values $\mathbf{s}_{i}$. The upper bound is decided by $\sum_{i=1}^{N}\mathbf{s}_{i}{-}\mathbf{s}_{1}$. For the clean and unperturbed feature, the score bound depends on $\sum_{i=1}^{N}\mathbf{s}_{i}$. Thus, removing the rank-1 feature with a larger $\mathbf{s}_{1}$ would reduce the upper bound of \texttt{RankFeat} score more. Considering the fact that OOD feature usually has a much larger $\mathbf{s}_{1}$ (see Fig.~\ref{fig:cover}(a)), \texttt{RankFeat} \textcolor{red}{is likely to} reduce the upper bound for OOD samples more. 
\end{proof}

%Our bound explicitly indicates the impact of $\mathbf{s}_{1}$ on the score. 

\noindent\textbf{Remark:} Considering that OOD feature usually has a much larger $\mathbf{s}_{1}$ (see Fig.~\ref{fig:cover}(a)), \texttt{RankFeat} would reduce the upper bound of OOD samples more and amplify the upper bound gap between ID and OOD data. 

\emph{Notice that our bound analysis strives to improve the understanding of OOD methods from new perspectives instead of giving a strict guarantee of the score.} For example, the upper bound can be used to explain the shrinkage and skew of score distributions in Fig.~\ref{fig:score_dist}. Subtracting $\mathbf{s}_{1}$ would largely reduce the numerical range of both ID and OOD scores, which could squeeze score distributions. Since the dominant singular value $\mathbf{s}_{1}$ contributes most to the score, removing $\mathbf{s}_{1}$ is likely to make many samples have similar scores. This would concentrate samples in a smaller region and further skew the distribution. Given that the OOD feature tends to have a much larger $\mathbf{s}_{1}$, this would have a greater impact on OOD data and skew the OOD score distribution more.

\textbf{Remark:} Though we use the assumption of GAP in Prop.~\ref{prop:up_bound}, this assumption is not really necessary for deriving our upper bounds. For example, when the model is Vision Transformer which does not have any pooling layer, the upper bounds still hold — the only difference is that the pooling vector $m$ will be removed in the bounds:
\begin{equation}
    \begin{aligned}
\texttt{RankFeat}(\mathbf{x}) &{<} \sum_{i=2}^{N} \mathbf{s}_{i} ||\mathbf{W}\mathbf{u}_{i}\mathbf{v}_{i}^{T}\cancel{\mathbf{m}}||_{\infty} {+} ||\mathbf{b}||_{\infty} {+} \log(Q)\\
&{\leq} \sum_{i=2}^{N} \mathbf{s}_{i} ||\mathbf{W}\cancel{\mathbf{m}}||_{\infty} {+}  ||\mathbf{b}||_{\infty} {+} \log(Q)\\
&{=}\cancel{\frac{1}{HW}} \left(\sum_{i=1}^{N} \mathbf{s}_{i} {-} \mathbf{s}_{1}\right) ||\mathbf{W}||_{\infty} {+}  ||\mathbf{b}||_{\infty} {+} \log(Q)
\end{aligned}
\end{equation}
As indicated above, the upper bounds of our \texttt{RankFeat} hold for general neural networks with fully connected layers.

\begin{figure}[tbp]
    \centering
    \includegraphics[width=0.99\linewidth]{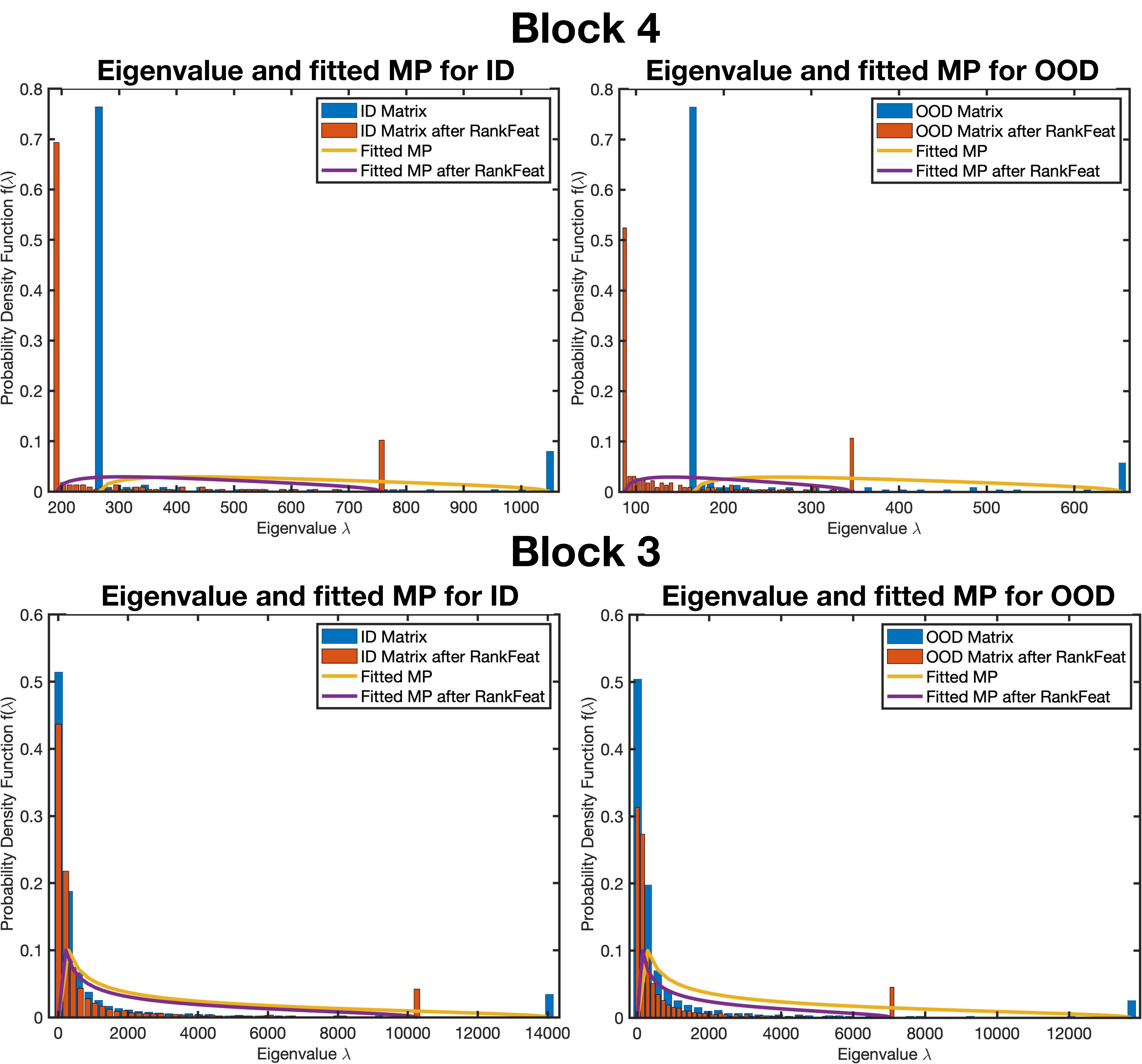}
    \caption{The exemplary eigenvalue distribution of ID/OOD feature and the fitted MP distribution. After the rank-1 matrix is removed, the lowest bin of OOD feature has a larger reduction and the middle bins gain some growth, making the ODD feature statistics closer to the MP distribution. }
    \label{fig:mp_dist}
\end{figure}

\subsection{Removing Rank-1 Matrix Makes the Statistics of OOD Features Closer to Random Matrices} 

Now we turn to use RMT to analyze the statistics of OOD and ID feature matrices. For a random matrix of a given shape, the density of its eigenvalue asymptotically converges to the Manchenko-Pastur (MP) distribution~\cite{marvcenko1967distribution,sengupta1999distributions}. Formally, we have the following theorem:
\begin{thm}[Manchenko-Pastur Law~\cite{marvcenko1967distribution,sengupta1999distributions}]
Let $\mathbf{X}$ be a random matrix of shape $t{\times}n$ whose entries are random variables with $E(\mathbf{X}_{ij}=0)$ and $E(\mathbf{X}_{ij}^2=1)$. Then the eigenvalues of the sample covariance $\mathbf{Y}=\frac{1}{n}\mathbf{X}\mathbf{X}^{T}$ converges to the probability density function: $\rho(\lambda) = \frac{t}{n} \frac{\sqrt{(\lambda_{+}-\lambda)(\lambda-\lambda_{-})}}{2\pi\lambda\sigma^2}\ for\  \lambda\in[\lambda_{-},\lambda_{+}]$ where $\lambda_{-}{=}\sigma^{2} (1-\sqrt{\frac{n}{t}})^2$ and $ \lambda_{+}{=}\sigma^{2} (1+\sqrt{\frac{n}{t}})^2$.
\end{thm}

We leave the proof in Sec. E of the Supplementary Material. This theorem implies the possibility of measuring the statistical distance between ID/OOD features and random matrices. To compute the statistical distance, we randomly sample $1,000$ ID and OOD feature matrices and compute the KL divergence between the actual eigenvalue distribution and the fitted MP distribution.

%Main Results
\begin{table*}[htbp]
    \caption{Main results on ResNetv2-101~\cite{he2016identity}. All values are reported in percentages, and these \emph{post hoc} methods are directly applied to the model pre-trained on ImageNet-1k~\cite{deng2009imagenet}. The best four results are highlighted with \textbf{\textcolor{red}{red}}, \textbf{\textcolor{blue}{blue}}, \textbf{\textcolor{cyan}{cyan}}, and \textbf{\textcolor{brown}{brown}}.}
    \centering
    \resizebox{0.99\linewidth}{!}{
    \begin{tabular}{c|cc|cc|cc|cc|cc}
    \toprule
        \multirow{3}*{\textbf{Methods}} & \multicolumn{2}{c|}{\textbf{iNaturalist}} & \multicolumn{2}{c|}{\textbf{SUN}} & \multicolumn{2}{c|}{\textbf{Places}} & \multicolumn{2}{c|}{\textbf{Textures}} & \multicolumn{2}{c}{\textbf{Average}}   \\
        \cmidrule{2-11}
         & FPR95 & AUROC  & FPR95 & AUROC  & FPR95 & AUROC  & FPR95 & AUROC & FPR95 & AUROC  \\
         & ($\downarrow$) & ($\uparrow$) & ($\downarrow$) & ($\uparrow$) & ($\downarrow$) & ($\uparrow$) & ($\downarrow$) & ($\uparrow$) & ($\downarrow$) & ($\uparrow$)\\
    \midrule
    MSP~\cite{hendrycks2016baseline} & 63.69 & 87.59 & 79.89 & 78.34 & 81.44 & 76.76 & 82.73 & 74.45 & 76.96 & 79.29\\
    ODIN~\cite{liang2017enhancing} & 62.69 & 89.36 & 71.67 & 83.92 & 76.27 & 80.67 & 81.31 & 76.30 & 72.99 & 82.56\\
    Energy~\cite{liu2020energy} & 64.91 & 88.48 & 65.33 & 85.32 & 73.02 & 81.37 & 80.87 & 75.79 & 71.03 & 82.74\\
    Mahalanobis~\cite{lee2018simple} &96.34 &46.33 &88.43 & 65.20 &89.75 &64.46 &52.23 &72.10 &81.69 &62.02\\
    GradNorm~\cite{huang2021importance} &50.03 &90.33 &46.48 &89.03 &60.86 &84.82 &61.42 &81.07 &54.70 &86.71\\
    ReAct~\cite{sun2021react} &\textbf{\textcolor{cyan}{44.52}}&\textbf{\textcolor{brown}{91.81}}&52.71&90.16&62.66&87.83&70.73&76.85&57.66&86.67\\
    %Igeood~\cite{gomes2022igeood} & & & & & & & & & &\\
    \midrule
    \rowcolor{gray!20}\textbf{RankFeat (Block 4)} &46.54 &81.49 &\textbf{\textcolor{blue}{27.88}} & \textbf{\textcolor{cyan}{92.18}} &\textbf{\textcolor{blue}{38.26}} & \textbf{\textcolor{cyan}{88.34}} & \textbf{\textcolor{brown}{46.06}} & \textbf{\textcolor{brown}{89.33}} & \textbf{\textcolor{cyan}{39.69}} &87.84\\
    \rowcolor{gray!20}\textbf{RankFeat (Block 3)} &49.61 &91.42&\textbf{\textcolor{brown}{39.91}} &\textbf{\textcolor{brown}{92.01}} &\textbf{\textcolor{brown}{51.82}} &\textbf{\textcolor{brown}{88.32}} &\textbf{\textcolor{cyan}{41.84}} &\textbf{\textcolor{cyan}{91.44}} &\textbf{\textcolor{brown}{45.80}} &\textbf{\textcolor{cyan}{90.80}} \\
     %\rowcolor{gray!20} \textbf{RankFeat (Score Average)}
     %&45.50&91.54&33.52&93.12&45.48&89.56&35.73&92.85&40.06&91.77\\
     \rowcolor{gray!20} \textbf{RankFeat (Block 3 + 4)}
     &\textbf{\textcolor{blue}{41.31}}&\textbf{\textcolor{cyan}{91.91}}&\textbf{\textcolor{cyan}{29.27}}&\textbf{\textcolor{blue}{94.07}}&\textbf{\textcolor{cyan}{39.34}}&\textbf{\textcolor{blue}{90.93}}&\textbf{\textcolor{blue}{37.29}}&\textbf{\textcolor{blue}{91.70}}&\textbf{\textcolor{blue}{36.80}}&\textbf{\textcolor{blue}{92.15}}\\
     %\rowcolor{gray!20} \textbf{RankFeat (Feature Average)}
     %&47.57&90.59&36.08&92.43&46.18&89.26&49.46&87.88&44.82&90.04\\
     %\rowcolor{gray!20} \textbf{GradNorm + ReAct} &50.03&90.33&46.48&89.03&60.86&84.82&61.42&81.07&&\\
     %\rowcolor{gray!20} \textbf{GradNorm + RankFeat (Block 4)} &50.03&90.33&46.48&89.03&&&&&&\\
     %\rowcolor{gray!20} \textbf{GradNorm + RankFeat (Block 3)} &50.03&90.33&&&&&&&&\\
     %\rowcolor{gray!20} \textbf{ReAct + RankFeat (Block 4)} &39.71&87.38&31.26&92.63&41.48&89.41&50.66&87.71&&\\
     %\rowcolor{gray!20} \textbf{ReAct + RankFeat (Block 3)} &36.99&93.51&31.69&93.57&42.89&90.21&&&&\\
     \midrule
     \rowcolor{gray!20} \textbf{RankWeight} &\textbf{\textcolor{brown}{45.96}}&\textbf{\textcolor{blue}{92.29}}&45.51&90.89&52.12&88.23&68.28&81.54&52.95&\textbf{\textcolor{brown}{88.24}}\\
     \rowcolor{gray!20} \textbf{RankFeat}+\textbf{RankWeight} &\textbf{\textcolor{red}{12.97}} &\textbf{\textcolor{red}{96.01}}&\textbf{\textcolor{red}{10.64}}&\textbf{\textcolor{red}{97.46}}&\textbf{\textcolor{red}{15.64}}&\textbf{\textcolor{red}{96.30}}&\textbf{\textcolor{red}{25.27}}&\textbf{\textcolor{red}{95.01}}&\textbf{\textcolor{red}{16.13}}&\textbf{\textcolor{red}{96.20}}\\
    \bottomrule
    \end{tabular}
    }
    \label{tab:main_results_res101}
\end{table*}
%According to Manchenko-Pastur (MP) law~\cite{marvcenko1967distribution,sengupta1999distributions}, given the shape $(t{\times}n)$ and variance $\sigma^2$ of a random matrix, the probability density function (pdf) of its eigenvalues asymptotically converges to the MP distribution
%\begin{equation}
%    \rho(\lambda) = \frac{t}{n} \frac{\sqrt{(\lambda_{+}-\lambda)(\lambda-\lambda_{-})}}{2\pi\lambda\sigma^2}\ if\  \lambda\in[\lambda_{-},\lambda_{+}] 
%\end{equation}
%where $\rho(\lambda){=}0$ for $\lambda$ outsides $[\lambda_{-},\lambda_{+}]$, and the eigenvalues bounds are $\lambda_{-}{=}\sigma^{2} (1-\sqrt{\frac{n}{t}})^2, \lambda_{+}{=}\sigma^{2} (1+\sqrt{\frac{n}{t}})^2$. To evaluate the statistical distance between ID/OOD matrix and random matrices, we randomly sample $1,000$ ID and OOD feature matrices. Then the KL divergence between the actual eigenvalue distribution and the fitted MP distribution is computed to measure the distance. 
%\begin{equation}
    %\lambda_{-} = \sigma^{2} (1-\sqrt{\frac{N}{T}})^2,\  \lambda_{+} = \sigma^{2} (1+\sqrt{\frac{N}{T}})^2
%\end{equation}
% Now we turn to using the RMT to analyse the singular value distribution of the ID and OOD feature.

\begin{table}[htbp]
    \caption{The KL divergence between ID/OOD feature and the fitted MP distribution. When the rank-1 feature or the rank-1 parameter is removed, the statistics of OOD matrix are closer to random matrices. }
    \centering
    \resizebox{0.99\linewidth}{!}{
    \begin{tabular}{c|cc|cc}
    \toprule
        \multirow{2}*{\textbf{Matrix Type}} & \multicolumn{2}{c|}{\textbf{Block 4}} & \multicolumn{2}{c}{\textbf{Block 3}} \\
        \cmidrule{2-5}
         & ID & OOD & ID & OOD \\
         \midrule
         Original feature matrix & 18.36 & 18.24 & 11.27 & 11.18\\
         Removing rank-1 feature & 17.07 ($\downarrow$ 1.29)& \textbf{15.79 ($\downarrow$ 2.45)}& 9.84 ($\downarrow$ 1.45) & \textbf{8.71 ($\downarrow$ 2.47)} \\
         Removing rank-1 weight &15.34 ($\downarrow$ 3.02) &\textbf{13.98 ($\downarrow$ 4.26)}  &8.52 ($\downarrow$ 2.75) &\textbf{6.79 ($\downarrow$ 4.39)}\\
    \bottomrule
    \end{tabular}
    }
    \label{tab:mp_distance}
\end{table}

Fig.~\ref{fig:mp_dist} and Table~\ref{tab:mp_distance} present the exemplary eigenvalue distribution and the average evaluation results of Block 4 and Block 3 features, respectively. For the original feature, the OOD and ID feature matrices exhibit similar behaviors: the distances to the fitted MP distribution are roughly the same ($diff.{\approx}0.1$). However, when we perform \texttt{RankFeat} or \texttt{RankWeight}, \emph{i.e.,} removing the rank-1 feature or parameter, the OOD feature matrix has a much larger drop in the KL divergence. This indicates that removing the rank-1 matrix makes the statistics of the OOD feature closer to random matrices, \emph{i.e.,} the OOD feature is very likely to become less informative than the ID feature. The result partly explains the working mechanism of \texttt{RankFeat} and \texttt{RankWeight}: \textit{by removing the feature matrix or the parameter matrix where OOD data might convey more information than ID data, the two distributions have a larger discrepancy and can be better separated.}

%This might imply that the rank-1 matrix of the OOD feature conveys more information than that of ID feature. Fig.~\ref{fig:mp_dist} displays the exemplary eigenvalue distribution and the corresponding MP distribution

\begin{table*}[htbp]
    \caption{The evaluation results on SqueezeNet~\cite{iandola2016squeezenet}, T2T-ViT-24~\cite{yuan2021tokens}, ViT-B/16~\cite{dosovitskiy2020vit}, and Swin-B~\cite{liu2021Swin}. All values are reported in percentages, and these \emph{post hoc} methods are directly applied to the model pre-trained on ImageNet-1k~\cite{deng2009imagenet}. The best four results are highlighted with \textbf{\textcolor{red}{red}}, \textbf{\textcolor{blue}{blue}}, \textbf{\textcolor{cyan}{cyan}}, and \textbf{\textcolor{brown}{brown}}. On Swin-B~\cite{liu2021Swin}, since the \texttt{Energy} score does not achieve reasonable performance, we select \texttt{MSP} as the base method for integrating our \texttt{RankFeat} and \texttt{RankWeight}.}
    \centering
    \resizebox{0.99\linewidth}{!}{
    \begin{tabular}{c|c|cc|cc|cc|cc|cc}
    \toprule
        \multirow{3}*{\textbf{Model}}&\multirow{3}*{\textbf{Methods}} & \multicolumn{2}{c|}{\textbf{iNaturalist}} & \multicolumn{2}{c|}{\textbf{SUN}} & \multicolumn{2}{c|}{\textbf{Places}} & \multicolumn{2}{c|}{\textbf{Textures}} & \multicolumn{2}{c}{\textbf{Average}}   \\
        \cmidrule{3-12}
         && FPR95 & AUROC  & FPR95 & AUROC  & FPR95 & AUROC  & FPR95 & AUROC & FPR95 & AUROC  \\
         && ($\downarrow$) & ($\uparrow$) & ($\downarrow$) & ($\uparrow$) & ($\downarrow$) & ($\uparrow$) & ($\downarrow$) & ($\uparrow$) & ($\downarrow$) & ($\uparrow$)\\
    \midrule
    \multirow{11}*{\textbf{SqueezeNet~\cite{iandola2016squeezenet}}}&MSP~\cite{hendrycks2016baseline} 
    &89.83&65.41&83.03&72.25&87.27&67.00&94.61&41.84&88.84&61.63\\
    &ODIN~\cite{liang2017enhancing} &90.79&65.75&78.32&78.37&83.23&73.31&92.25&43.43&86.15&65.17\\
    &Energy~\cite{liu2020energy} &79.27&73.30&56.41&87.88&67.74&82.73&67.16&64.51&67.65&77.11\\
    &Mahalanobis~\cite{lee2018simple} &91.50&51.79&90.33&62.18&92.26&56.63&58.60&67.16&83.17&59.44\\
    &GradNorm~\cite{huang2021importance} &76.31&73.92&53.63&87.55&65.99&83.28&68.72&68.07&66.16&78.21\\
    &ReAct~\cite{sun2021react} &76.78&68.56&87.57&66.37&88.80&66.20&51.05&76.57&76.05&69.43\\
    %Igeood~\cite{gomes2022igeood} & & & & & & & & & &\\
    \cmidrule{2-12}
   \rowcolor{gray!20} \cellcolor{white}&\textbf{RankFeat (Block 4)} &\textbf{\textcolor{blue}{61.67}}&\textbf{\textcolor{blue}{83.09}}&\textbf{\textcolor{brown}{46.72}}&88.31&\textbf{\textcolor{cyan}{61.31}}&{80.52}&\textbf{\textcolor{blue}{38.04}}&\textbf{\textcolor{blue}{88.82}}&\textbf{\textcolor{blue}{51.94}}&\textbf{\textcolor{cyan}{85.19}}\\
    \rowcolor{gray!20} \cellcolor{white}&\textbf{RankFeat (Block 3)} &\textbf{\textcolor{brown}{71.04}}&\textbf{\textcolor{brown}{81.50}}&49.18&\textbf{\textcolor{blue}{90.43}}&62.94&\textbf{\textcolor{red}{85.82}}&\textbf{\textcolor{brown}{50.14}}&\textbf{\textcolor{brown}{79.32}}&\textbf{\textcolor{brown}{58.33}}&\textbf{\textcolor{brown}{84.28}} \\
    \rowcolor{gray!20} \cellcolor{white}&\textbf{RankFeat (Block 3 + 4)}
    &\textbf{\textcolor{cyan}{65.81}}&\textbf{\textcolor{cyan}{83.06}}&\textbf{\textcolor{cyan}{46.64}}&\textbf{\textcolor{cyan}{90.17}}&\textbf{\textcolor{brown}{61.56}}&\textbf{\textcolor{cyan}{84.51}}&\textbf{\textcolor{cyan}{42.54}}&\textbf{\textcolor{cyan}{85.00}}&\textbf{\textcolor{cyan}{54.14}}&\textbf{\textcolor{blue}{85.69}}\\
    \cmidrule{2-12}
    \rowcolor{gray!20}\cellcolor{white}&\textbf{RankWeight}&77.91&62.77&\textbf{\textcolor{blue}{44.03}}&\textbf{\textcolor{brown}{89.89}}&\textbf{\textcolor{blue}{58.35}}&\textbf{\textcolor{brown}{84.11}}&61.75&66.26&60.51&75.76 \\
    \rowcolor{gray!20}\cellcolor{white}&\textbf{RankFeat}+\textbf{RankWeight} &\textbf{\textcolor{red}{53.18}}&\textbf{\textcolor{red}{87.46}}&\textbf{\textcolor{red}{39.21}}&\textbf{\textcolor{red}{91.55}}&\textbf{\textcolor{red}{54.88}}&\textbf{\textcolor{blue}{85.39}}&\textbf{\textcolor{red}{33.89}}&\textbf{\textcolor{red}{91.07}}&\textbf{\textcolor{red}{45.29}}&\textbf{\textcolor{red}{88.87}}\\
    \midrule
    \multirow{9}*{\textbf{T2T-ViT-24~\cite{yuan2021tokens}}}&MSP~\cite{hendrycks2016baseline} &48.92&\textbf{\textcolor{brown}{88.95}}&\textbf{\textcolor{brown}{61.77}}&\textbf{\textcolor{cyan}{81.37}}&69.54&\textbf{\textcolor{cyan}{80.03}}&62.91&82.31&60.79&83.17\\ 
    &ODIN~\cite{liang2017enhancing} &\textbf{\textcolor{cyan}{44.07}}&88.17&63.83&78.46&\textbf{\textcolor{brown}{68.19}}&75.33&54.27&83.63&\textbf{\textcolor{brown}{57.59}}&81.40\\
    &Energy~\cite{liu2020energy} &52.95&82.93&68.55&73.06&74.24&68.17&\textbf{\textcolor{brown}{51.05}}&83.25&61.70&76.85\\
    &Mahalanobis~\cite{lee2018simple} &90.50&58.13&91.71&50.52&93.32&49.60&80.67&64.06&89.05&55.58\\
    &GradNorm~\cite{huang2021importance} &99.30&25.86&98.37&28.06&99.01&25.71&92.68&38.80&97.34&29.61\\
    &ReAct~\cite{sun2021react} &52.17&\textbf{\textcolor{cyan}{89.51}}&65.23&\textbf{\textcolor{brown}{81.03}}&68.93&\textbf{\textcolor{brown}{78.20}}&52.54&\textbf{\textcolor{brown}{85.46}}&59.72&\textbf{\textcolor{brown}{83.55}}\\
    \cmidrule{2-12}
   %\rowcolor{gray!20} \cellcolor{white}&\textbf{RankFeat} &\textbf{41.17}&\textbf{90.53}&\textbf{54.70}&\textbf{85.94}&\textbf{62.46}&\textbf{83.18}&\textbf{34.10}&\textbf{89.27}&\textbf{48.10}&\textbf{87.23}\\
   \rowcolor{gray!20} \cellcolor{white}&\textbf{RankFeat} &\textbf{\textcolor{brown}{50.27}} & 87.81 & \textbf{\textcolor{blue}{57.18}} & \textbf{\textcolor{blue}{84.33}} & \textbf{\textcolor{cyan}{66.22}} & \textbf{\textcolor{blue}{80.89}} & \textbf{\textcolor{blue}{32.64}} & \textbf{\textcolor{blue}{89.36}} &\textbf{\textcolor{cyan}{51.58}} & \textbf{\textcolor{blue}{85.60}} \\
   \cmidrule{2-12}
    \rowcolor{gray!20}\cellcolor{white}&\textbf{RankWeight} &\textbf{\textcolor{blue}{22.54}}&\textbf{\textcolor{blue}{94.27}}&\textbf{\textcolor{cyan}{61.76}}&79.86&\textbf{\textcolor{blue}{63.28}}&77.39&\textbf{\textcolor{cyan}{44.11}}&\textbf{\textcolor{cyan}{86.66}}&\textbf{\textcolor{blue}{47.92}}&\textbf{\textcolor{cyan}{84.55}}\\
    %\rowcolor{gray!20}\cellcolor{white}&\textbf{RankFeat}+\textbf{RankWeight} &44.50&90.37&52.45&87.23&60.77&84.23&28.49&90.43&46.55&88.07\\
    \rowcolor{gray!20}\cellcolor{white}&\textbf{RankFeat}+\textbf{RankWeight} &\textbf{\textcolor{red}{20.36}}&\textbf{\textcolor{red}{95.52}}&\textbf{\textcolor{red}{47.33}}&\textbf{\textcolor{red}{88.48}}&\textbf{\textcolor{red}{55.39}}&\textbf{\textcolor{red}{85.74}}&\textbf{\textcolor{red}{27.48}}&\textbf{\textcolor{red}{90.65}}&\textbf{\textcolor{red}{37.64}}&\textbf{\textcolor{red}{90.10}}\\
    \midrule
    % \multirow{2}*{\textbf{DeiT-B16~\cite{touvron2021training}}} &MSP~\cite{hendrycks2016baseline} &50.51 &88.95	&71.14 &78.93 &72.08 &78.63	&81.72 &65.40 &68.86 &77.98\\
    % &ODIN~\cite{liang2017enhancing} &50.23 &88.12 &69.71 &77.22	&71.37 &76.26 &82.00 &60.95 &68.33 &75.64 \\
    % \midrule
    \multirow{9}*{\textbf{ViT-B/16~\cite{dosovitskiy2020vit}}} &MSP~\cite{hendrycks2016baseline} &\textbf{\textcolor{blue}{47.96}} &\textbf{\textcolor{blue}{88.57}}	&\textbf{\textcolor{brown}{69.55}} &\textbf{\textcolor{cyan}{78.95}}	&\textbf{\textcolor{brown}{71.15}} &\textbf{\textcolor{cyan}{77.96}}	&63.76 &\textbf{\textcolor{cyan}{79.84}} &\textbf{\textcolor{brown}{63.11}} &\textbf{\textcolor{brown}{81.33}}\\
    &ODIN~\cite{liang2017enhancing} &\textbf{\textcolor{red}{46.13}} &\textbf{\textcolor{brown}{87.33}} &\textbf{\textcolor{cyan}{68.70}} &\textbf{\textcolor{brown}{74.67}} &\textbf{\textcolor{cyan}{70.68}} &\textbf{\textcolor{brown}{72.65}} &\textbf{\textcolor{cyan}{62.11}} &77.02 &\textbf{\textcolor{cyan}{61.91}} &\textbf{\textcolor{cyan}{77.92}}\\
    &Energy~\cite{liu2020energy} &63.30 &80.89 &76.80 &68.00 &77.87 &65.47 &67.06 &73.76 &71.26 &72.03\\
    &Mahalanobis~\cite{lee2018simple} &98.95 &25.49 &93.18 &56.67 &92.56 &58.07 &89.50 &54.70 &93.55 &48.73\\
    &GradNorm~\cite{huang2021importance} &89.92 &54.41 &96.02 &41.89 &96.30 &39.61 &91.79 &47.04 &93.51 &45.74\\
    &ReAct~\cite{sun2021react} &72.99 &81.77 &79.29 &73.42 &79.22 &71.72 &67.98 &\textbf{\textcolor{brown}{77.85}} &74.87 &76.19\\
    \cmidrule{2-12}
    \rowcolor{gray!20} \cellcolor{white}&\textbf{RankFeat} &\textbf{\textcolor{brown}{62.90}} &\textbf{\textcolor{cyan}{88.18}} &\textbf{\textcolor{blue}{61.51}} &\textbf{\textcolor{blue}{83.53}} &\textbf{\textcolor{blue}{63.14}} &\textbf{\textcolor{blue}{81.51}} &\textbf{\textcolor{blue}{55.46}} &\textbf{\textcolor{blue}{85.35}} &\textbf{\textcolor{blue}{60.75}} &\textbf{\textcolor{blue}{84.64}}\\
    \cmidrule{2-12}
    \rowcolor{gray!20}\cellcolor{white}&\textbf{RankWeight} &70.45 &77.09 &78.31 &68.05 &79.19 &66.36 &\textbf{\textcolor{brown}{62.80}} &77.68 &72.69 &72.30\\
    \rowcolor{gray!20}\cellcolor{white}&\textbf{RankFeat}+\textbf{RankWeight} &\textbf{\textcolor{cyan}{48.93}} &\textbf{\textcolor{red}{90.27}} &\textbf{\textcolor{red}{51.85}} &\textbf{\textcolor{red}{86.58}} &\textbf{\textcolor{red}{53.33}} &\textbf{\textcolor{red}{84.81}} &\textbf{\textcolor{red}{49.68}} &\textbf{\textcolor{red}{87.81}} &\textbf{\textcolor{red}{50.95}} &\textbf{\textcolor{red}{87.37}}\\
    \midrule
    \multirow{9}*{\textbf{Swin-B~\cite{liu2021Swin}}} &MSP~\cite{hendrycks2016baseline} &\textbf{\textcolor{brown}{52.81}} &\textbf{\textcolor{brown}{85.06}} &\textbf{\textcolor{brown}{71.50}} &\textbf{\textcolor{brown}{76.48}} &\textbf{\textcolor{brown}{71.35}} &\textbf{\textcolor{brown}{77.11}} &\textbf{\textcolor{brown}{76.54}} &\textbf{\textcolor{brown}{67.16}} &\textbf{\textcolor{brown}{68.05}} &\textbf{\textcolor{brown}{76.45}}\\
    &ODIN~\cite{liang2017enhancing} & 58.06 &81.49 &72.22 &72.73 &71.82 &73.74 &79.26 &61.63 &70.34 &72.40\\
    &Energy~\cite{liu2020energy} &86.85 &65.09 &83.54 &63.38 &82.80 &64.86 &89.13 &51.31 &85.58 &61.16\\
    &Mahalanobis~\cite{lee2018simple} &94.92 &63.64 &95.44 &46.11 &94.18 &47.26 &79.68 &65.63 &91.06 &55.66\\
    &GradNorm~\cite{huang2021importance} &97.43 &31.61 &94.08 &36.89 &94.02 &38.73 &95.48 &30.37 &95.25 &34.40\\
    &ReAct~\cite{sun2021react} &80.93 &77.32 &79.32 &72.05 &78.96 &72.71 &83.90 &63.64 &80.78 &71.43\\
    \cmidrule{2-12}
    \rowcolor{gray!20} \cellcolor{white}&\textbf{RankFeat}* & \textbf{\textcolor{cyan}{51.38}} &\textbf{\textcolor{cyan}{87.66}} &\textbf{\textcolor{blue}{60.32}} &\textbf{\textcolor{blue}{84.52}} &\textbf{\textcolor{blue}{64.77}} &\textbf{\textcolor{blue}{82.29}} &\textbf{\textcolor{blue}{72.36}} &\textbf{\textcolor{blue}{81.04}} &\textbf{\textcolor{blue}{62.21}} &\textbf{\textcolor{blue}{83.88}}\\
    %\rowcolor{gray!20} \cellcolor{white}&\textbf{RankFeat+Energy} & 97.67 &57.63 &89.97 &68.55 &89.48 &67.74 &80.96 &76.30 &89.52 &67.56\\
    \cmidrule{2-12}
    \rowcolor{gray!20}\cellcolor{white}&\textbf{RankWeight}* &\textbf{\textcolor{blue}{49.14}} &\textbf{\textcolor{blue}{87.89}} &\textbf{\textcolor{cyan}{67.24}} &\textbf{\textcolor{cyan}{81.18}} &\textbf{\textcolor{cyan}{67.75}} &\textbf{\textcolor{cyan}{81.68}} &\textbf{\textcolor{cyan}{74.04}} &\textbf{\textcolor{cyan}{71.90}} &\textbf{\textcolor{cyan}{64.54}} &\textbf{\textcolor{cyan}{80.66}}\\
    %\rowcolor{gray!20}\cellcolor{white}&\textbf{RankWeight+Energy} &79.78 &74.75 &75.2 &73.72 &74.55 &74.95 &84.08 &60.79 &78.4 &71.05\\
    \rowcolor{gray!20}\cellcolor{white}&\textbf{RankFeat}+\textbf{RankWeight}* &\textbf{\textcolor{red}{39.64}} &\textbf{\textcolor{red}{90.32}} &\textbf{\textcolor{red}{58.47}} &\textbf{\textcolor{red}{85.57}} &\textbf{\textcolor{red}{62.87}} &\textbf{\textcolor{red}{82.89}} &\textbf{\textcolor{red}{48.28}} &\textbf{\textcolor{red}{87.72}} &\textbf{\textcolor{red}{52.32}} &\textbf{\textcolor{red}{86.63}}\\
    %\rowcolor{gray!20}\cellcolor{white}&\textbf{RankFeat}+\textbf{RankWeight}+Energy &89.85 &71.91 &84.14 &75.74 &85.96 &72.92 &58.92 &84.07 &79.72 &76.16\\
    \bottomrule
    \end{tabular}
    }
    \label{tab:main_results_vit}
\end{table*}

\subsection{Connection between RankFeat and ReAct~\cite{sun2021react}}

\texttt{ReAct} clips the activations at the penultimate layer of a model to distinguish ID and OOD samples. Given the feature $\mathbf{X}$ and the pooling layer $\mathbf{m}$, the perturbation can be defined as:
\begin{equation}
     \min(\mathbf{X}\mathbf{m},\tau) = \mathbf{X}\mathbf{m} - \max(\mathbf{X}\mathbf{m}-\tau,0) 
\end{equation}
where $\tau$ is a pre-defined threshold. Their method shares some similarity with \texttt{RankFeat} formulation $\mathbf{X}\mathbf{m}{-}\mathbf{s}_{1}\mathbf{u}_{1}\mathbf{v}_{1}^{T}\mathbf{m}$. \textit{Both approaches subtract from the feature a portion of information that is most likely to cause the over-confidence of OOD prediction.} \texttt{ReAct} selects the manually-defined threshold $\tau$ based on statistics of the whole ID set, while \texttt{RankFeat} generates the structured rank-1 matrix from the feature itself. Taking a step further, \texttt{ReAct} has the score inequality following eq.~\eqref{eq:score_upper}
\begin{equation}
\begin{gathered}
    \texttt{ReAct}(\mathbf{x})<  || \mathbf{W}\mathbf{X}\mathbf{m} - \mathbf{W}\max(\mathbf{X}\mathbf{m}-\tau,0) ||_{\infty}\\ + ||\mathbf{b}||_{\infty} + \log(Q)
\end{gathered}
\end{equation}
Since $\mathbf{X}$ is non-negative (output of \texttt{ReLU}),
we have $\max(\mathbf{X}\mathbf{m})\geq\nicefrac{\max(\mathbf{X})}{HW}$. Exploiting the vector norm inequality $||\mathbf{X}||_{\rm F}{\geq}||\mathbf{X}||_{2}$ leads to the relation $\max(\mathbf{X}){\geq}\nicefrac{\mathbf{s}_{1}}{\sqrt{CHW}}$. Relying on this property, the above inequality can be re-formulated as:
\begin{equation}
\begin{gathered}
    \texttt{ReAct}(\mathbf{x})< \frac{1}{HW}\sum_{i=1}^{N} \mathbf{s}_{i} ||\mathbf{W}||_{\infty}\\ - \boxed{\frac{1}{HW}{\max\left( \frac{\mathbf{s}_{1} }{\sqrt{CHW}}-\tau, 0 \right)||\mathbf{W}||_{\infty}}}+ ||\mathbf{b}||_{\infty} + \log(Q)
\end{gathered}
\end{equation}
As indicated above, the upper bound of \texttt{ReAct} is also determined by the largest singular value $\mathbf{s}_{1}$. In contrast, the upper bound of our \texttt{RankFeat} can be expressed as:
\begin{equation}
\begin{gathered}
    \texttt{RankFeat}(\mathbf{x})< \frac{1}{HW}\sum_{i=1}^{N} \mathbf{s}_{i} ||\mathbf{W}||_{\infty}\\ - \boxed{\frac{1}{HW}\mathbf{s}_{1} ||\mathbf{W}||_{\infty}} + ||\mathbf{b}||_{\infty} + \log(Q)
\end{gathered}
\end{equation}
The upper bounds of both methods resemble each other with the only different term boxed. \textit{From this point of view, both methods distinguish the ID and OOD data by eliminating the impact of the term containing $\mathbf{s}_{1}$ in the upper bound.} \texttt{ReAct} optimizes it by clipping the term with a manually-defined threshold, which is indirect and might be sub-optimal. Moreover, the threshold selection requires statistics of the whole ID set. In contrast, our \texttt{RankFeat} does not require any extra data and directly subtracts this underlying term which is likely to cause the over-confidence of OOD samples. 

\begin{figure*}[t]
    \centering
    \includegraphics[width=0.99\linewidth]{imgs/rankweight_vra_ash.pdf}
    \caption{Score distributions of some OOD detection~\cite{djurisic2023extremely,xu2023vra} approached integrated with our \texttt{RankWeight} on ResNetv2-101.}
    \label{fig:rankweight_vra_ash}
\end{figure*}

\subsection{RankWeight Tightens the Upper Bounds}

Given the parameter matrix $\mathbf{M}$ and the feature $\mathbf{X}_{L-1}$ before this layer, we generate the target high-level feature as:
\begin{equation}
    \mathbf{X}_L = \mathbf{M} \mathbf{X}_{L-1}
\end{equation}
where $\mathbf{X}_L$ corresponds to the feature $\mathbf{X}$ in the bound analysis of Eq.~(\ref{eq:bound_x}), and the bias is neglected for simplicity. The above equation holds for linear layers in Transformers and also convolution layers in CNNs (convolutional kernels can be extended to circular Toeplitz matrices for matrix multiplication). When our \texttt{RankWeight} is performed, the feature would become:
\begin{equation}
    \mathbf{X}_L' = \mathbf{M}'\mathbf{X}_{L-1} = \mathbf{M}\mathbf{X}_{L-1} - \Tilde{\mathbf{s}}_{1}\Tilde{\mathbf{u}}_{1}\Tilde{\mathbf{v}}_{1}^{T}\mathbf{X}_{L-1}
\end{equation}
Let $\Tilde{\mathbf{s}}_{2}$ denotes the second largest singular value of $\mathbf{M}$. Then according to the inequality $\sigma_i(\mathbf{A}\mathbf{B})\leq\sigma_{max}(\mathbf{A})\sigma_i(\mathbf{B})$, the singular values of the new feature would satisfy:
\begin{equation}
    \sigma_i(\mathbf{X}_L') \leq\Tilde{\mathbf{s}}_{2}\sigma_i(\mathbf{X}_{L-1})
\end{equation}
where $\sigma_i(\cdot)$ denote the $i$-th singular value. Notice that for the original feature $\mathbf{X}_L$, the singular values have:
\begin{equation}
    \mathbf{s}_i = \sigma_i(\mathbf{X}_L)\leq \Tilde{\mathbf{s}}_{1} \sigma_i(\mathbf{X}_{L-1})
\end{equation}
We can write the upper bound of \texttt{RankWeight} as:
\begin{equation}
\begin{aligned}
     \texttt{RankWeight}(\mathbf{x}) &\leq \frac{\sum_{i=1}^{N} \sigma_i(\mathbf{X}_L')}{HW}||\mathbf{W}||_{\infty} + C \\
     &\leq \frac{\sum_{i=1}^{N} \Tilde{\mathbf{s}}_{2} \sigma_i(\mathbf{X}_{L-1})}{HW}||\mathbf{W}||_{\infty} + C\\
     &=\frac{\Tilde{\mathbf{s}}_{2}}{\Tilde{\mathbf{s}}_{1}} \frac{\sum_{i=1}^{N} \Tilde{\mathbf{s}}_{1} \sigma_i(\mathbf{X}_{L-1})}{HW}||\mathbf{W}||_{\infty} + C
\end{aligned}
\end{equation}
Here we use $C$ to represent the constants $||\mathbf{b}||_{\infty} + \log(Q)$. Recall that for the \texttt{Energy} score, the upper bound is:
\begin{equation}
    \texttt{Energy}(\mathbf{x}) \leq  \frac{\sum_{i=1}^{N} \Tilde{\mathbf{s}}_{1} \sigma_i(\mathbf{X}_{L-1})}{HW}||\mathbf{W}||_{\infty} + C
\end{equation}
Since we have $\nicefrac{\Tilde{\mathbf{s}}_{2}}{\Tilde{\mathbf{s}}_{1}}\leq1$, it is clear that our \texttt{RankWeight} tightens the upper bound of the \texttt{Energy} score. Similarly, when \texttt{RankWeight} is combined with \texttt{RankFeat} or \texttt{ReAct}, we see the upper bound is also scaled by a factor of $\nicefrac{\Tilde{\mathbf{s}}_{2}}{\Tilde{\mathbf{s}}_{1}}\leq1$. Specifically, for \texttt{RankFeat} we have:
\begin{equation}
\begin{gathered}
    \texttt{RankWeight}(\mathbf{x}) + \texttt{RankFeat}(\mathbf{x})  {\leq} 
     \frac{\Tilde{\mathbf{s}}_{2}}{\Tilde{\mathbf{s}}_{1}} \frac{1}{HW} \left(\sum_{i=1}^{N} \mathbf{s}_{i} - \mathbf{s}_{1}\right)\\ ||\mathbf{W}||_{\infty} +  ||\mathbf{b}||_{\infty} + \log(Q)
\end{gathered}
\end{equation}
For \texttt{ReAct}, we have:
\begin{equation}
\begin{gathered}
    \texttt{RankWeight}(\mathbf{x}) + \texttt{ReAct}(\mathbf{x})  {\leq} 
     \frac{\Tilde{\mathbf{s}}_{2}}{\Tilde{\mathbf{s}}_{1}} \frac{1}{HW}\Big(\sum_{i=1}^{N} \mathbf{s}_{i}\\ - {\max\left( \frac{\mathbf{s}_{1} }{\sqrt{CHW}}-\tau, 0 \right) \Big)||\mathbf{W}||_{\infty}} + ||\mathbf{b}||_{\infty} + \log(Q)
\end{gathered}
\end{equation}
In a nutshell, our \texttt{RankWeight} re-scales the singular values of the upper bound, thus making the bound tighter and the associated analysis more practically useful. 

%RankWeight global perturbation, while RankFeat local disturb

%Moreover, when \texttt{RankFeat} is jointly used, our \texttt{RankWeight} further tightens the upper bound of \texttt{RankFeat}.

%In a nutshell, RankWeight global perturbation, while RankFeat local disturb
\begin{table*}[t]
    \caption{Comparisons on ResNetv2-101~\cite{he2016identity} of some OOD detection approaches empowered by our \texttt{RankWeight}. Our \texttt{RankWeight} brings about consistent performance gain on all the testsets for every baseline.}
    \centering
    \resizebox{0.99\linewidth}{!}{
    \begin{tabular}{c|cc|cc|cc|cc|cc}
    \toprule
        \multirow{3}*{\textbf{Methods}} & \multicolumn{2}{c|}{\textbf{iNaturalist}} & \multicolumn{2}{c|}{\textbf{SUN}} & \multicolumn{2}{c|}{\textbf{Places}} & \multicolumn{2}{c|}{\textbf{Textures}} & \multicolumn{2}{c}{\textbf{Average}}   \\
        \cmidrule{2-11}
         & FPR95 & AUROC  & FPR95 & AUROC  & FPR95 & AUROC  & FPR95 & AUROC & FPR95 & AUROC  \\
         & ($\downarrow$) & ($\uparrow$) & ($\downarrow$) & ($\uparrow$) & ($\downarrow$) & ($\uparrow$) & ($\downarrow$) & ($\uparrow$) & ($\downarrow$) & ($\uparrow$)\\
    \midrule
    GradNorm~\cite{huang2021importance} &50.03 &90.33 &46.48 &89.03 &60.86 &84.82 &61.42 &81.07 &54.70 &86.71\\
    GradNorm~\cite{huang2021importance} + RankWeight& \textbf{15.53}&\textbf{97.44}&\textbf{20.58} &\textbf{96.55}&\textbf{26.26}&\textbf{95.53}&\textbf{57.57}&\textbf{86.39}&\textbf{29.99} ($\uparrow$ \textbf{24.71}) & \textbf{93.93}($\uparrow$ \textbf{7.22})\\
    \midrule
    ReAct~\cite{sun2021react} &44.52&91.81&52.71&90.16&62.66&87.83&70.73&76.85&57.66&86.67\\
    ReAct~\cite{sun2021react} + RankWeight& \textbf{20.42}&\textbf{96.53}&\textbf{31.99}&\textbf{94.84}&\textbf{37.67}&\textbf{93.37}&\textbf{54.40}&\textbf{84.75}&\textbf{36.12} ($\uparrow$ \textbf{21.54})&\textbf{92.37} ($\uparrow$ \textbf{5.70}) \\
    \midrule
    VRA~\cite{xu2023vra} & 20.81&97.70&32.89&92.68&45.83&90.01&23.88&95.43 &30.85&93.71\\
    VRA~\cite{xu2023vra}+ RankWeight&\textbf{8.90} &\textbf{98.37}&\textbf{15.64} &\textbf{96.99}&\textbf{21.62} &\textbf{95.85}&\textbf{24.82}&\textbf{95.16}&\textbf{17.75} ($\uparrow$ \textbf{13.10})&\textbf{96.59} ($\uparrow$ \textbf{2.88})  \\
    \midrule
    ASH~\cite{djurisic2023extremely} &21.22&96.37&38.42&90.85&51.37&88.10&14.73&97.19&31.44&93.13 \\
    ASH~\cite{djurisic2023extremely}+ RankWeight&\textbf{3.60}&\textbf{99.07}&\textbf{19.14}&\textbf{96.45}&\textbf{26.19}&\textbf{95.46}&\textbf{12.89}&\textbf{97.57}&\textbf{15.46} ($\uparrow$ \textbf{15.98})&\textbf{97.13} ($\uparrow$ \textbf{4.00}) \\
    \bottomrule
    \end{tabular}
    }
    \label{tab:rankweight}
\end{table*}

\section{Experimental Results}
\label{sec:exp}

In this section, we first discuss the setup in Sec.~\ref{sec:exp_setup}, and then present the main large-scale experimental results in Sec.~\ref{sec:exp_result}, 
followed by the extensive ablation studies in Sec.~\ref{sec:exp_ablation}.

\subsection{Setup}
\label{sec:exp_setup}

\noindent \textbf{Datasets.} In line with~\cite{huang2021mos,sun2021react,huang2021importance}, we mainly evaluate our method on the large-scale ImageNet-1k benchmark~\cite{deng2009imagenet}. The large-scale dataset is more challenging than the traditional CIFAR benchmark~\cite{krizhevsky2009learning} because the images are more realistic and diverse (\emph{i.e.,} $1.28$M images of $1,000$ classes). For the OOD datasets, we select four testsets from subsets of \texttt{iNaturalist}~\cite{van2018inaturalist}, \texttt{SUN}~\cite{xiao2010sun}, \texttt{Places}~\cite{zhou2017places}, and \texttt{Textures}~\cite{cimpoi2014describing}. These datasets are crafted by~\cite{huang2021mos} with non-overlapping categories from ImageNet-1k. 
Besides these four OOD test sets, we further evaluate our method on the large-scale Species dataset. The Species~\cite{hendrycks2019scaling} dataset is a large-scale OOD validation benchmark consisting of $71,3449$ images, which is designed for ImageNet-1k~\cite{deng2009imagenet} and ImageNet 21-k~\cite{kolesnikov2020big} as the ID sets. We select four sub-sets as the OOD benchmark, namely \texttt{Protozoa}, \texttt{Microorganisms}, \texttt{Plants}, and \texttt{Mollusks}. 
Besides the experiment on the large-scale benchmark, we also validate the effectiveness of our approach on the CIFAR~\cite{krizhevsky2009learning} benchmark (see Sec. A of the Supplementary Material).

%For more details about the used datasets, please refer to the supplementary material.

\noindent \textbf{Baselines.}
We compare our method with $6$ recent \emph{post hoc} OOD detection methods, namely \texttt{MSP}~\cite{hendrycks2016baseline}, \texttt{ODIN}~\cite{liang2017enhancing}, \texttt{Energy}~\cite{liu2020energy}, \texttt{Mahalanobis}~\cite{lee2018simple}, \texttt{GradNorm}~\cite{huang2021importance}, and \texttt{ReAct}~\cite{sun2021react}. Also, we show that the two recent activation-shaping based approaches, \emph{i.e.,} \texttt{ASH}~\cite{djurisic2023extremely} and \texttt{VRA}~\cite{xu2023vra}, can be further boosted by our \texttt{RankWeight}. The detailed illustration and settings are kindly referred to Sec. B of the Supplementary Material.

\noindent \textbf{Architectures.} In line with~\cite{huang2021importance}, the main evaluation is done using Google BiT-S model~\cite{kolesnikov2020big} pretrained on ImageNet-1k with ResNetv2-101~\cite{he2016identity}. We also evaluate the performance on SqueezeNet~\cite{iandola2016squeezenet}, an alternative tiny architecture suitable for mobile devices, and on T2T-ViT-24~\cite{yuan2021tokens}, a tokens-to-tokens vision transformer that has impressive performance when trained from scratch. We also benchmark our method on two other popular transformer architectures, \emph{i.e.,} Vision Transformers (ViTs)~\cite{dosovitskiy2020vit} and SwinTransformer~\cite{liu2021Swin}.

%\textit{For the implementation details and evaluation metrics, please refer to Supplementary Material}.

\noindent \textbf{Implementation Details.} At the inference stage, all the images are resized to $480{\times}480$ for ResNetv2-101~\cite{he2016identity} and SqueezeNet~\cite{iandola2016squeezenet}. The source codes are implemented with \texttt{Pytorch 1.10.1}, and all experiments are run on a single NVIDIA Quadro RTX 6000 GPU.

\noindent \textbf{Evaluation Metrics.} Following~\cite{huang2021mos,sun2021react,huang2021importance}, we measure the performance using two main metrics: (1) the false positive rate (FPR95) of OOD examples when the true positive rate of ID samples is at 95\%; and (2) the area under the receiver operating characteristic curve (AUROC). 

\begin{figure}[htbp]
    \centering
\begin{lstlisting}[language=Python]
#Our RankFeat (SVD) is applied on each individual \\
#feature matrix within the mini-batch.
feat = model.features(inputs)
B, C, H, W = feat.size()
feat = feat.view(B, C, H * W)
u,s,vt = torch.linalg.svd(feat)
feat = feat - s[:,0:1].unsqueeze(2)*u[:,:,0:1].bmm(vt[:,0:1,:])
feat = feat.view(B,C,H,W)
logits = model.classifier(feat)
score = torch.logsumexp(logits, dim=1)
\end{lstlisting}
    \caption{Pytorch-like codes of our \texttt{RankFeat} implementation.}
    \label{fig:code_feat}
\end{figure}

\noindent \textbf{Pseudo Code.} Fig.~\ref{fig:code_feat} and~\ref{fig:code_weight} present the Pytorch-like implementation of our \texttt{RankFeat} and \texttt{RankWeight}, respectively. We use \texttt{torch.linalg.svd} to perform the SVD.

\subsection{Results}
\label{sec:exp_result}

\begin{table*}[t]
    \caption{The evaluation results on four sub-sets of Species~\cite{hendrycks2019scaling} based on ResNetv2-101~\cite{he2016identity}. All values are reported in percentages, and these \emph{post hoc} methods are directly applied to the model pre-trained on ImageNet-1k~\cite{deng2009imagenet}. The best four results are highlighted with \textbf{\textcolor{red}{red}}, \textbf{\textcolor{blue}{blue}}, \textbf{\textcolor{cyan}{cyan}}, and \textbf{\textcolor{brown}{brown}}.}
    \centering
    \resizebox{0.99\linewidth}{!}{
    \begin{tabular}{c|cc|cc|cc|cc|cc}
    \toprule
        \multirow{3}*{\textbf{Methods}} & \multicolumn{2}{c|}{\textbf{Protozoa}} & \multicolumn{2}{c|}{\textbf{Microorganisms}} & \multicolumn{2}{c|}{\textbf{Plants}} & \multicolumn{2}{c|}{\textbf{Mollusks}} & \multicolumn{2}{c}{\textbf{Average}}   \\
        \cmidrule{2-11}
         & FPR95 & AUROC  & FPR95 & AUROC  & FPR95 & AUROC  & FPR95 & AUROC & FPR95 & AUROC  \\
         & ($\downarrow$) & ($\uparrow$) & ($\downarrow$) & ($\uparrow$) & ($\downarrow$) & ($\uparrow$) & ($\downarrow$) & ($\uparrow$) & ($\downarrow$) & ($\uparrow$)\\
    \midrule
    MSP~\cite{hendrycks2016baseline} &75.81&83.20&72.23&84.25& 61.48 & 87.78 &85.62&70.51 & 73.79 & 81.44 \\
    ODIN~\cite{liang2017enhancing} &75.97&85.11&65.94&89.35&55.69&90.79  &86.22&71.31 & 70.96&84.14\\
    Energy~\cite{liu2020energy} &79.49&84.34 &60.87&\textbf{\textcolor{brown}{90.30}} &54.67&90.95&88.47&70.53&70.88&84.03 \\
    %Mahalanobis~\cite{lee2018simple} &96.34 &46.33 &88.43 & 65.20 &89.75 &64.46 &52.23 &72.10 &81.69 &62.02\\
    %GradNorm~\cite{huang2021importance} &50.03 &90.33 &46.48 &89.03 &60.86 &84.82 &61.42 &81.07 &54.70 &86.71\\
    ReAct~\cite{sun2021react} &81.74&84.26&58.82&85.88&\textbf{\textcolor{cyan}{36.90}}&\textbf{\textcolor{cyan}{93.78}}&90.58&76.33&67.02&85.06\\
    \midrule
    \rowcolor{gray!20}\textbf{RankFeat (Block 4)} & 66.98&70.19&\textbf{\textcolor{brown}{39.06}}&86.67&46.31&79.98&\textbf{\textcolor{brown}{80.14}}&59.92&\textbf{\textcolor{brown}{58.12}}&74.19\\
    \rowcolor{gray!20}\textbf{RankFeat (Block 3)} & \textbf{\textcolor{brown}{58.99}}&\textbf{\textcolor{cyan}{88.81}}&49.72&90.04&47.01&91.85&80.37&\textbf{\textcolor{cyan}{79.61}}&59.02&\textbf{\textcolor{brown}{87.58}}\\
     \rowcolor{gray!20} \textbf{RankFeat (Block 3 + 4)}& \textbf{\textcolor{cyan}{52.78}}&\textbf{\textcolor{brown}{88.65}}&\textbf{\textcolor{cyan}{37.21}}&\textbf{\textcolor{cyan}{92.82}}&\textbf{\textcolor{brown}{38.07}}&\textbf{\textcolor{brown}{92.88}}&\textbf{\textcolor{cyan}{76.38}}&\textbf{\textcolor{brown}{78.13}}&\textbf{\textcolor{cyan}{51.11}}&\textbf{\textcolor{cyan}{88.37}}\\
     \midrule
      \rowcolor{gray!20} \textbf{RankWeight} & \textbf{\textcolor{blue}{47.49}}&\textbf{\textcolor{blue}{92.10}}&\textbf{\textcolor{blue}{29.94}}&\textbf{\textcolor{blue}{95.23}}&\textbf{\textcolor{blue}{24.76}}&\textbf{\textcolor{blue}{95.79}}&\textbf{\textcolor{blue}{65.32}}&\textbf{\textcolor{blue}{84.32}}&\textbf{\textcolor{blue}{41.88}}&\textbf{\textcolor{blue}{91.86}}\\
      \rowcolor{gray!20} \textbf{RankFeat} + \textbf{RankWeight} &\textbf{\textcolor{red}{27.57}}&\textbf{\textcolor{red}{92.55}}&\textbf{\textcolor{red}{14.17}}&\textbf{\textcolor{red}{96.19}}&\textbf{\textcolor{red}{9.51}}&\textbf{\textcolor{red}{97.67}}&\textbf{\textcolor{red}{37.94}}&\textbf{\textcolor{red}{86.58}}&\textbf{\textcolor{red}{22.30}}&\textbf{\textcolor{red}{93.25}} \\
    \bottomrule
    \end{tabular}
    }
    \label{tab:species_res101}
\end{table*}

\begin{figure}[htbp]
    \centering
\begin{lstlisting}[language=Python]
weight = model.block4[-1].conv3.weight.data
u,s,vt  = torch.linalg.svd(weight)
weight = weight - s[:,0:1].unsqueeze(2)*u[:,:,0:1].bmm(vt[:,0:1,:])
model.block4[-1].conv3.weight.data = weight
\end{lstlisting}
    \caption{Pytorch-like codes of our \texttt{RankWeight} core implementation. Here we prune the last parameter matrix before the fully-connected layer by removing the rank-1 weight. }
    \label{fig:code_weight}
\end{figure}

\subsubsection{Main Results on ImageNet-1k}

\begin{table}[htbp]
    \centering
    \caption{Comparison against training-needed methods on ImageNet-1k based on ResNetv2-101~\cite{he2016identity}.}
    \resizebox{0.99\linewidth}{!}{
    \begin{tabular}{c|c|c|c|c}
    \toprule
         Method & Post hoc? & Free of Validation Set? & FPR95 ($\downarrow$) & AUROC ($\uparrow$)\\
    \midrule
         KL Matching~\cite{hendrycks2019scaling}& $\textcolor{green}{\cmark}$ &$\textcolor{red}{\xmark}$&54.30&80.82 \\
         MOS~\cite{huang2021mos} & $\textcolor{red}{\xmark}$ & $\textcolor{green}{\cmark}$ &39.97&90.11 \\
         \rowcolor{gray!20}\textbf{RankFeat} & $\textcolor{green}{\cmark}$ & $\textcolor{green}{\cmark}$ & 36.80 & 92.15\\
         %KL Matching~\cite{hendrycks2019scaling}& $\textcolor{green}{\checkmark}$ &$\textcolor{red}{\pmb\times}$&54.30&80.82 \\
         %MOS~\cite{huang2021mos} & $\textcolor{red}{\pmb\times}$ & $\textcolor{green}{\checkmark}$ &39.97&90.11 \\
         %\rowcolor{gray!20}\textbf{RankFeat} & $\textcolor{green}{\checkmark}$ & $\textcolor{green}{\checkmark}$ & \textbf{36.80} &\textbf{92.15}\\
         \rowcolor{gray!20}\textbf{RankWeight} & $\textcolor{green}{\cmark}$ & $\textcolor{green}{\cmark}$ &52.95 & 88.24\\
         \rowcolor{gray!20}\textbf{RankFeat}+\textbf{RankWeight} & $\textcolor{green}{\cmark}$ & $\textcolor{green}{\cmark}$ &\textbf{16.13} &\textbf{96.20} \\
    \bottomrule
    \end{tabular}
    }
    \label{tab:mos_kl}
\end{table}

Following~\cite{huang2021importance}, the main evaluation is conducted using Google BiT-S model~\cite{kolesnikov2020big} pretrained on ImageNet-1k with ResNetv2-101 architecture~\cite{he2016identity}. Table~\ref{tab:main_results_res101} compares the performance of all the \emph{post hoc} methods. For both Block 3 and Block 4 features, our \texttt{RankFeat} achieves impressive evaluation results across datasets and metrics. More specifically, \texttt{RankFeat} based on the Block 4 feature outperforms the previous best baseline by $\textbf{15.01\%}$ in the average FPR95, while the Block 3 feature-based \texttt{RankFeat} beats the previous best method by $\textbf{4.09\%}$ in the average AUROC. Their combination further surpasses other methods by $\textbf{17.90\%}$ in the average FPR95 and by $\textbf{5.44\%}$ in the average AUROC. The superior performances at various depths demonstrate the effectiveness and general applicability of \texttt{RankFeat}. The Block 3 feature has a higher AUROC but slightly falls behind the Block 4 feature in the FPR95, which can be considered a compromise between the two metrics. For our \texttt{RankWeight}, the evaluation results of its standalone usage are very competitive against previous baselines. Moreover, jointly using \texttt{RankFeat} and \texttt{RankWeight} establishes the new \emph{state-of-the-art} of this benchmark, achieving a low FPR95 of $\textbf{16.13\%}$ and a high AUROC of $\textbf{96.20\%}$. This implies that our \texttt{RankFeat} and \texttt{RankWeight} can complement each other to maximally separate and distinguish the ID and OOD data. 

%The Block 3 feature has a higher AUROC but the FPR95 score is slightly inferior to that of Block 4 feature, which can be considered as a compromise between the two metrics. This also indicates that these high-level features capture rich semantic information and the over-confident prediction of OOD samples are closely related to the rank-1 matrix of the high-level feature. 

\subsubsection{RankFeat/RankWeight Works on Alternative CNNs}

Besides the experiment on ResNetv2~\cite{he2016identity}, we also evaluate our method on SqueezeNet~\cite{iandola2016squeezenet}, an alternative tiny network suitable for mobile devices and on-chip applications. This network is more challenging because the tiny network size makes the model prone to overfit the training data, which could increase the difficulty of distinguishing between ID and OOD samples. Table~\ref{tab:main_results_vit} top presents the performance of all the methods. Collectively, the performance of \texttt{RankFeat} and \texttt{RankWeight} is very competitive. Specifically, our \texttt{RankFeat} outperforms the previous best baseline by $\textbf{14.22\%}$ in FPR95 and by $\textbf{7.48\%}$ in AUROC, while our \texttt{RankWeight} surpasses the previous best method by $\textbf{5.65\%}$ in FPR95. When \texttt{RankFeat} and \texttt{RankWeight} are used simultaneously, the results achieve \emph{state-of-the-art} performance, attaining a low FPR95 of $\textbf{45.29\%}$ and a high AUROC of \textbf{88.87\%}.

%a popular transformer architecture that can achieve competitive performance against CNNs when trained from scratch. 

\subsubsection{RankFeat/RankWeight also Suits Transformers} 

To further demonstrate the applicability of our method, we evaluate \texttt{RankFeat} and \texttt{RankWeight} on Tokens-to-Tokens Vision Transformer (T2T-ViT)~\cite{yuan2021tokens}, Vision Transformers (ViTs)~\cite{dosovitskiy2020vit}, and SwinTransformer~\cite{liu2021Swin}. Similar to the CNNs, our methods are performed on the final and penultimate tokens of T2T-ViT and ViT, respectively. Due to the sliding $7{\times}7$ window, the feature maps of SwinTransformer are small in dimensionality at the last layer; we hence perform our \texttt{RankFeat} and \texttt{RankWeight} on the third-to-last layer. Table~\ref{tab:main_results_vit} bottom compares the performance on T2T-ViT-24, ViT-B/16, and Swin-B. Our \texttt{RankFeat} outperforms the second-best method by $\textbf{2.05\%}$ in AUROC on T2T-ViT-24, by $\textbf{3.31\%}$ in AUROC on ViT-B/16, and by $\textbf{7.43\%}$ in AUROC on Swin-B. Since the transformer models~\cite{dosovitskiy2020image,yuan2021tokens} do not have increasing receptive fields like CNNs, we do not evaluate the performance at alternative network depths. When our \texttt{RankWeight} is applied, the FPR95 further reduces to $\textbf{37.64\%}$ and the AUROC increases to $\textbf{90.10\%}$ on T2T-ViT-24, the FPR95 further reduces to $\textbf{50.95\%}$ and the AUROC increases to $\textbf{87.37\%}$ on ViT-B/16, and the FPR95 further reduces to $\textbf{52.32\%}$ and the AUROC increases to $\textbf{86.63\%}$ on Swin-B. These experimental results indicate the flexible composability of our \texttt{RankFeat} and \texttt{RankWeight} on various Transformers. 

\begin{table}[htbp]
    \caption{Ablation studies on keeping only the rank-1 matrix and removing the rank-n matrix. }
    \centering
    \resizebox{0.99\linewidth}{!}{
    \begin{tabular}{c|cc|cc|cc|cc|cc}
    \toprule
        \multirow{3}*{\textbf{Baselines}} & \multicolumn{2}{c|}{\textbf{iNaturalist}} & \multicolumn{2}{c|}{\textbf{SUN}} & \multicolumn{2}{c|}{\textbf{Places}} & \multicolumn{2}{c|}{\textbf{Textures}} & \multicolumn{2}{c}{\textbf{Average}}   \\
        \cmidrule{2-11}
         & FPR95 & AUROC  & FPR95 & AUROC  & FPR95 & AUROC  & FPR95 & AUROC & FPR95 & AUROC  \\
         & ($\downarrow$) & ($\uparrow$) & ($\downarrow$) & ($\uparrow$) & ($\downarrow$) & ($\uparrow$) & ($\downarrow$) & ($\uparrow$) & ($\downarrow$) & ($\uparrow$)\\
    \midrule
    GradNorm~\cite{huang2021importance} &50.03 &90.33 &46.48 &89.03 &60.86 &84.82 &61.42 &81.07 &54.70 &86.71\\
    ReAct~\cite{sun2021react} &\textbf{44.52}&{91.81}&52.71&90.16&62.66&87.83&70.73&76.85&57.66&86.67\\
    \midrule
    Keeping Only Rank-1 &48.97&\textbf{91.93}&62.63&84.62&72.42&79.79&49.42&88.86&58.49&86.30 \\
    Removing Rank-3 &55.19&90.03&48.97&91.26&56.63&\textbf{88.81}&86.95&74.57&61.94&86.17\\
    Removing Rank-2 &50.04&89.30&48.55&90.99&56.23&88.38&76.86&81.37&57.92&87.51\\
    \midrule
    \rowcolor{gray!20}\textbf{Removing Rank-1} &\textbf{46.54} &81.49 &\textbf{27.88} & \textbf{92.18} &\textbf{38.26} & {88.34} & \textbf{46.06} & \textbf{89.33} & \textbf{39.69} &\textbf{87.84}\\
    \bottomrule
    \end{tabular}
    }
    \label{tab:ablation_rank}
\end{table}

\subsubsection{Comparison against Training-needed Approaches} 

Since our method is \emph{post hoc}, we only compare it with other \emph{post hoc} baselines. MOS~\cite{huang2021mos} and KL Matching~\cite{hendrycks2019scaling} are not taken into account because MOS needs extra training processes and KL Matching requires the labeled validation set to compute distributions for each class. Nonetheless, we note that our method can still hold an advantage against those approaches. Table~\ref{tab:mos_kl} presents the average FPR95 and AUROC of these methods on the ImageNet-1k benchmark. Without any prior knowledge of the training or validation ID set, our \texttt{RankFeat} as well as its combination with \texttt{RankWeight} achieve better performance than previous training-needed approaches, outperforming previous methods by $\textbf{23.84\%}$ in average FPR95 and $\textbf{6.09\%}$. 
 
%Our \texttt{RankFeat} achieves the best performance without any extra training or validation set.

%\subsubsection{Comparison against activation-shaping methods}

\begin{table}[htbp]
    \caption{Ablation studies on applying \texttt{RankFeat} to features at different network depths.} %based on ResNetv2-101. The Block 4 and Block 3 features are the most informative.}
    \centering
    \resizebox{0.99\linewidth}{!}{
    \begin{tabular}{c|cc|cc|cc|cc|cc}
    \toprule
        \multirow{3}*{\textbf{Layer}} & \multicolumn{2}{c|}{\textbf{iNaturalist}} & \multicolumn{2}{c|}{\textbf{SUN}} & \multicolumn{2}{c|}{\textbf{Places}} & \multicolumn{2}{c|}{\textbf{Textures}} & \multicolumn{2}{c}{\textbf{Average}}   \\
        \cmidrule{2-11}
         & FPR95 & AUROC  & FPR95 & AUROC  & FPR95 & AUROC  & FPR95 & AUROC & FPR95 & AUROC  \\
         & ($\downarrow$) & ($\uparrow$) & ($\downarrow$) & ($\uparrow$) & ($\downarrow$) & ($\uparrow$) & ($\downarrow$) & ($\uparrow$) & ($\downarrow$) & ($\uparrow$)\\
    \midrule
    Block 1 &87.81&77.00&59.15&87.29&65.50&84.35&94.15&60.41&76.65&77.26\\
    Block 2 &71.84&85.80&61.44&86.46&71.68&81.65&87.89&72.04&73.23&81.49\\
    \rowcolor{gray!20}\textbf{Block 3} &\textbf{49.61} &\textbf{91.42} &\textbf{39.91} &\textbf{92.01} &\textbf{51.82} &\textbf{88.32} &\textbf{41.84} &\textbf{91.44} &\textbf{45.80} &\textbf{90.80} \\
    \rowcolor{gray!20}\textbf{Block 4} &\textbf{46.54} &81.49 &\textbf{27.88} & \textbf{92.18} &\textbf{38.26} & \textbf{88.34} & \textbf{46.06} & \textbf{89.33} & \textbf{39.69} &\textbf{87.84}\\
    \bottomrule
    \end{tabular}
    }
    \label{tab:ablation_block}
\end{table}

%As we discussed in Sec, there are some recent 

\subsubsection{RankWeight Enhances Other OOD Approaches}
\label{sec:rankweight_integration}

Table~\ref{tab:rankweight} shows the performance of some OOD detection approaches incorporated with our \texttt{RankWeight}. Empowered by our method, each baseline gets consistent performance gain across different datasets with an average improvement of $\textbf{18.83\%}$ in FPR95 and $\textbf{4.95\%}$ in AUROC. In particular, both \texttt{VRA} and \texttt{ASH} establish \emph{state-of-the-art} performance when integrated with our \texttt{RankWeight}. The compatibility with a wide range of baselines demonstrates that our \texttt{RankWeight} serves as a general add-on for OOD detection methods.

\subsubsection{Results on the Large-scale Species Dataset}

We present the evaluation results on Species in Table~\ref{tab:species_res101}. 
The results are coherent with our previous experiments on alternative benchmarks and architectures. Specifically, our \texttt{RankFeat} surpasses other methods by $\textbf{15.91\%}$ in the average FPR95 and by $\textbf{3.31\%}$ in the average AUROC, and our \texttt{RankWeight} outperforms other approaches by $\textbf{25.14\%}$ in FPR95 and by \textbf{6.80\%} in AUROC. Moreover, the joint use of \texttt{RankFeat} and \texttt{RankWeight} achieves the best performance, reducing FPR95 by \textbf{44.72\%} and improving AUROC by \textbf{8.19\%} compared with previous approaches. 

\subsection{Ablation Studies}
\label{sec:exp_ablation}

In this subsection, we conduct ablation studies based on ResNetv2-101. Unless explicitly specified, our \texttt{RankFeat} is applied to the Block 4 feature by default. 

\subsubsection{Removing the Rank-1 Matrix Outperforms Keeping it}

Instead of removing the rank-1 matrix, another seemingly promising approach is keeping only the rank-1 matrix and abandoning the rest of the matrix. Table~\ref{tab:ablation_rank} presents the evaluation results of keeping only the rank-1 matrix. The performance falls behind that of removing the rank-1 feature by $18.8\%$ in FPR95, which indicates that keeping only the rank-1 feature is inferior to removing it in distinguishing the two distributions. Nonetheless, it is worth noting that even keeping only the rank-1 matrix achieves very competitive performance against previous best methods, such as \texttt{GradNorm}~\cite{huang2021importance} and \texttt{ReAct}~\cite{sun2021react}.

\subsubsection{Removing the Rank-1 Matrix Outperforms Removing the Rank-n Matrix (n$>$1)} 

We evaluate the impact of removing the matrix of a higher rank, \emph{i.e.,} performing $\mathbf{X}{-}\sum_{i=1}^{n}\mathbf{s}_{i}\mathbf{u}_{i}\mathbf{v}_{i}^{T}$ where $n{>}1$ for the high-level feature $\mathbf{X}$. Table~\ref{tab:ablation_rank} compares the performance of removing the rank-2 matrix and rank-3 matrix. When the rank of the removed matrix is increased, the average performance degrades accordingly. This demonstrates that removing the rank-1 matrix is the most effective approach to separate ID and OOD data. This result is coherent with the finding in Fig.~\ref{fig:cover}(a): only the largest singular value of OOD data is significantly different from that of ID data. Therefore, removing the rank-1 matrix achieves the best performance.  

\begin{table}[htbp]
    \caption{The approximate solution by PI yields competitive performance and costs much less time consumption. The test batch size is set as $16$.}
    \centering
    \resizebox{0.99\linewidth}{!}{
    \begin{tabular}{c|c|cc|cc|cc|cc|cc}
    \toprule
        \multirow{3}*{\makecell[c]{\textbf{Computation} \\ \textbf{Technique}}} &
        \multirow{3}*{\makecell[c]{\textbf{Processing Time} \\ \textbf{Per Image} \\ \textbf{(ms)}}} 
        & \multicolumn{2}{c|}{\textbf{iNaturalist}} & \multicolumn{2}{c|}{\textbf{SUN}} & \multicolumn{2}{c|}{\textbf{Places}} & \multicolumn{2}{c|}{\textbf{Textures}} & \multicolumn{2}{c}{\textbf{Average}}   \\
        \cmidrule{3-12}
         & & FPR95 & AUROC  & FPR95 & AUROC  & FPR95 & AUROC  & FPR95 & AUROC & FPR95 & AUROC  \\
         & & ($\downarrow$) & ($\uparrow$) & ($\downarrow$) & ($\uparrow$) & ($\downarrow$) & ($\uparrow$) & ($\downarrow$) & ($\uparrow$) & ($\downarrow$) & ($\uparrow$)\\
         \midrule
          GradNorm~\cite{huang2021importance} &80.01
          &50.03 &90.33 &46.48 &89.03 &60.86 &84.82 &61.42 &81.07 &54.70 &86.71\\
          ReAct~\cite{sun2021react} &8.79 &\textbf{44.52}&\textbf{91.81}&52.71&90.16&62.66&87.83&70.73&76.85&57.66&86.67\\
          \midrule
         SVD & 18.01 &{46.54} &81.49 &\textbf{27.88} & \textbf{92.18} &{38.26} & \textbf{88.34} & \textbf{46.06} & \textbf{89.33} & \textbf{39.69} &\textbf{87.84}\\
         PI ($\#100$ iter) &9.97 &46.59&81.49&27.93&92.18&38.28&88.34&46.09&89.33&39.72&\textbf{87.84}\\
         PI ($\#50$ iter) &9.47 &46.58&81.49&27.93&92.17&\textbf{38.24}&88.34&46.12&89.32 &39.72&87.83\\
         PI ($\#20$ iter) &9.22 &46.58 &81.48 &27.93 &92.15 &38.28 &88.31 &46.10 &89.33 &39.75 &87.82 \\
         PI ($\#10$ iter) &9.03 &46.77&81.29&28.21&91.84&38.44&87.94&46.08&89.37&39.88&87.61 \\ 
         PI ($\#5$ iter) &9.00&48.34&79.81&30.44&89.71&41.33&84.97&45.34&89.41&41.36&85.98 \\ 
         %PI ($\#3$ iter) &8.92&52.75&76.21&40.36&82.98&50.02&78.11&44.10&88.72&46.86&81.51\\
         %PI ($\#1$ iter) &8.91&83.15&69.60&76.64&79.83&82.06&76.00&73.40&69.73&78.81&73.79\\
    \bottomrule
    \end{tabular}
    }
    \label{tab:ablation_approximate}
\end{table}

\subsubsection{Block 3 and Block 4 Features are the Most Informative}

In addition to exploring the high-level features at Block 3 and Block 4, we also investigate the possibility of applying \texttt{RankFeat} to features at shallow network layers. As shown in Table~\ref{tab:ablation_block}, the performances of \texttt{RankFeat} at the Block 1 and Block 2 features are not comparable to those at deeper layers. This is mainly because the shallow low-level features do not embed as rich semantic information as the deep features. Consequently, removing the rank-1 matrix of shallow features would not help to separate ID/OOD data. 

%High-level features capture more semantic information.

%Block4 feature 2048x225(15x15)
%Block3 feature 2048x900(30x30)

%\noindent \textbf{Effect of Temperature.}

\subsubsection{Approximate PI Yields Competitive Performance}
\label{sec:pi_com}

Table~\ref{tab:ablation_approximate} compares the time consumption and performance of SVD and PI, as well as two recent \emph{state-of-the-art} OOD methods \texttt{ReAct} and \texttt{GradNorm}. The performance of PI starts to become competitive against that of SVD (${<}0.1\%$) from $\textbf{20}$ iterations on with $\textbf{48.41\%}$ time reduction. Compared with \texttt{ReAct}, the PI-based \texttt{RankFeat} only requires marginally $4.89\%$ more time consumption. \texttt{GradNorm} is not comparable against other baselines in terms of time cost because it does not support the batch mode.

\begin{figure}[htbp]
    \centering
    \includegraphics[width=0.9\linewidth]{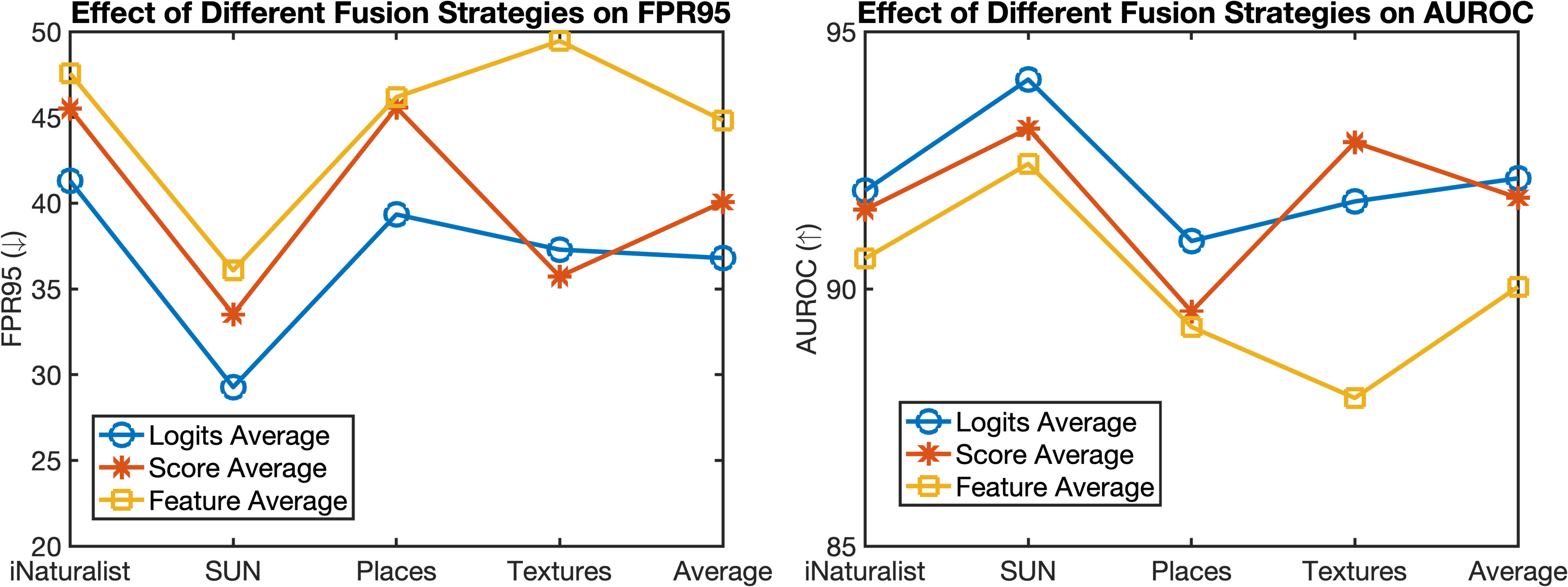}
    \caption{The impact of fusion strategies on FPR95 and AUROC.}
    \label{fig:fusion}
\end{figure}

\subsubsection{Logit Fusion Achieves the Best Performance} 
\label{sec:fusion}

Fig.~\ref{fig:fusion} displays the performance of different fusion strategies in combining \texttt{RankFeat} at the Block 3 and Block 4 features. As can be observed, averaging the logits outperforms other fusion strategies in most datasets and metrics. This indicates that fusing the logits can best coordinate and complement the benefit of both features at different depths.   

\begin{table}[htbp]
    \caption{Impact of \texttt{RankWeight} when pruning different numbers of successive layers before the fully-connected layer. }
    \centering
    \resizebox{0.99\linewidth}{!}{
    \begin{tabular}{c|cc|cc|cc|cc|cc}
    \toprule
        \multirow{3}*{\textbf{\# Layer}} & \multicolumn{2}{c|}{\textbf{iNaturalist}} & \multicolumn{2}{c|}{\textbf{SUN}} & \multicolumn{2}{c|}{\textbf{Places}} & \multicolumn{2}{c|}{\textbf{Textures}} & \multicolumn{2}{c}{\textbf{Average}}   \\
        \cmidrule{2-11}
         & FPR95 & AUROC  & FPR95 & AUROC  & FPR95 & AUROC  & FPR95 & AUROC & FPR95 & AUROC  \\
         & ($\downarrow$) & ($\uparrow$) & ($\downarrow$) & ($\uparrow$) & ($\downarrow$) & ($\uparrow$) & ($\downarrow$) & ($\uparrow$) & ($\downarrow$) & ($\uparrow$)\\
    \midrule
     \rowcolor{gray!20}1 &12.97&96.01&10.64&97.46&15.64&96.30&25.27&95.01&\textbf{16.13}&\textbf{96.20}\\
     2 &16.66&94.87&10.24&97.43&15.05&96.31&24.57&94.88&16.63&95.87\\
     3 &15.15&95.22&11.26&97.28&16.12&96.21&26.26&94.66&17.20&95.84\\   
    \bottomrule
    \end{tabular}
    }
    \label{tab:rankweight_layers}
\end{table}

\subsubsection{Pruning only One Layer is Sufficient}
\label{sec:pruning}

An interesting question about \texttt{RankWeight} is how many layers we need to prune. Table~\ref{tab:rankweight_layers} compares the performance of \texttt{RankWeight} when pruning more layers. We can observe that adding more layers does not really help in improving the performance, and pruning a single layer is sufficient. The importance of this layer also implies that the over-confidence of OOD samples is more likely to happen in the last parametric layer before the fully-connected layer. 

\subsubsection{Pruning the Last Layer is Most Effective}
\label{sec:pruning_depth}

\begin{table}[htbp]
    \caption{Impact of \texttt{RankWeight} when pruning weight matrices at different depths on ResNetv2-101.}
    \centering
    \resizebox{0.99\linewidth}{!}{
    \begin{tabular}{c|cc|cc|cc|cc|cc}
    \toprule
        \multirow{3}*{\textbf{Block}} & \multicolumn{2}{c|}{\textbf{iNaturalist}} & \multicolumn{2}{c|}{\textbf{SUN}} & \multicolumn{2}{c|}{\textbf{Places}} & \multicolumn{2}{c|}{\textbf{Textures}} & \multicolumn{2}{c}{\textbf{Average}}   \\
        \cmidrule{2-11}
         & FPR95 & AUROC  & FPR95 & AUROC  & FPR95 & AUROC  & FPR95 & AUROC & FPR95 & AUROC  \\
         & ($\downarrow$) & ($\uparrow$) & ($\downarrow$) & ($\uparrow$) & ($\downarrow$) & ($\uparrow$) & ($\downarrow$) & ($\uparrow$) & ($\downarrow$) & ($\uparrow$)\\
    \midrule
     \rowcolor{gray!20}4 &12.97&96.01&10.64&97.46&15.64&96.30&25.27&95.01&\textbf{16.13}&\textbf{96.20}\\
     3 &51.35&91.25&38.46&92.48&50.85&88.58&42.32&91.44&45.75&90.94\\
     2 &46.89&80.68&33.68&89.34&42.97&85.39&44.17&89.71&41.93&86.28\\
     1 &44.10&82.40&33.17&89.93&44.59&85.00&44.68&89.60&41.64&86.73\\
    \bottomrule
    \end{tabular}
    }
    \label{tab:rankweight_depth}
\end{table}

Table~\ref{tab:rankweight_depth} presents the evaluation results of applying \texttt{RankWeight} at different depths. As can be seen, performing \texttt{RankWeight} at earlier layers will greatly deteriorate the performance. This indicates that applying our \texttt{RankWeight} to prune the last layer is most effective. 

%GradNorm feature layer
%57.15 83.74|33.89 93.33|50.77 88.49|53.63 80.67|48.86 86.56
%59.79 83.16|37.91 87.24| -rank1
%63.92 79.50|34.15 92.83|51.10 87.43|60.28 78.11

%Block4 Rank-1 Weight Removal
%10.50 96.94|14.68 96.79|19.94 95.46|27.64 94.57|18.19 95.94 RankWeight+RankFeat
%45.96 92.29|45.51 90.89|52.12 88.23|68.28 81.54|52.95 88.24 RankWeight
%51.35 91.25|38.46 92.48|50.85 88.58|42.32 91.44| Block3
%15.15 95.22|11.26 97.28|16.12 96.21|26.26 94.66|17.20 95.84 Conv13
%16.66 94.87｜10.24 97.43｜15.05 96.31｜24.57 94.88｜16.63 95.87 Conv123
%12.97 96.01|10.64 97.46｜15.64 96.30｜25.27 95.01｜16.13 96.20 Conv23
%14.62 97.35|14.21 96.30|21.89 95.33|22.48 95.33| block3+4

%ResNet-50
%51.90 84.43|62.22 84.90|75.06 77.27|19.65 95.64|52.21 85.56 RankWeight+RankFeat
%40.12 93.26|73.53 84.13|74.60 82.66|76.88 77.69| RankWeight
%47.51 91.68|36.79 92.99|43.76 90.89|88.97 69.25| RankWeight+ReAct

%RankWeight 
%20.42 96.53|31.99 94.84|37.67 93.37|54.40 84.75| +React
%15.53 97.44|20.58 96.55|26.26 95.53|57.57 86.39| +GradNorm
%3.60  99.07|19.14 96.45|26.19 95.46|12.89 97.57|15.46 97.13 +ASH
%8.90  98.37|15.64 96.99|21.62 95.85|24.82 95.16|17.75 96.59 +VRA

%% file: 5_conclusion.tex
\section{Conclusion}
\label{sec:conc}
In this paper, we present \texttt{RankFeat} and \texttt{RankWeight}, two simple yet effective approaches for \emph{post hoc} OOD detection. \texttt{RankFeat} performs OOD detection by removing the rank-1 matrix composed of the largest singular value from the high-level feature, while \texttt{RankWeight} similarly removes the rank-1 matrix from the parameter matrices of only one deep layer. We demonstrate its superior empirical results and the general applicability across architectures, network depths, different composability, and various benchmarks. Extensive ablation studies and comprehensive theoretical analyses are conducted to reveal important insights and to explain the working mechanism of our method.

%% file: 6_supp.tex
%\section{Experimental Setup}
%\label{sec:exp}

\section{More Evaluation Results}

%\subsection{Comparison against MOS and KL Matching}

%Since our method is \emph{post hoc}, we only compare other \emph{post hoc} baselines in the paper. MOS~\cite{huang2021mos} and KL Matching~\cite{hendrycks2019scaling} are not taken into account for comparison because MOS needs extra training processes and KL Matching requires the labeled validation set to compute distributions for each class. However, our method can even hold an advantage over those approaches. Table~\ref{tab:mos_kl} compares the average FPR95 and AUROC on the ImageNet-1k benchmark. Our RankFeat achieves the best performance without any extra training or validation set.

%\begin{table}[htbp]
%    \centering
%    \caption{Comparison against training-needed methods on ImageNet-1k based on ResNetv2-101~\cite{he2016identity}.}
%    \begin{tabular}{c|c|c|c|c}
%    \toprule
%         Method & Post hoc? & Free of Validation Set? & FPR95 ($\downarrow$) & AUROC ($\uparrow$)\\
%    \midrule
%         KL Matching~\cite{hendrycks2019scaling}& $\textcolor{green}{\cmark}$ &$\textcolor{red}{\xmark}$&54.30&80.82 \\
%         MOS~\cite{huang2021mos} & $\textcolor{red}{\xmark}$ & $\textcolor{green}{\cmark}$ &39.97&90.11 \\
%         \rowcolor{gray!20}\textbf{RankFeat} & $\textcolor{green}{\cmark}$ & $\textcolor{green}{\cmark}$ & \textbf{36.80} &\textbf{92.15}\\
%    \bottomrule
%    \end{tabular}
%    \label{tab:mos_kl}
%\end{table}

\subsection{CIFAR100 with Different Architectures}
%To ensure the fair comparison and facilitate the reproducibility, we adopt ResNet-56~\cite{he2016deep} model and RepVGG-A0~\cite{ding2021repvgg} model pre-trained on CIFAR100 datast from the public published models via \textsc{torch.hub}\footnote{The pre-trained model is loaded from this public repository \href{https://github.com/chenyaofo/pytorch-cifar-models}{https://github.com/chenyaofo/pytorch-cifar-models} which is published via \textsc{torch.hub}.}.

\begin{table}[htbp]
    \centering 
    \caption{The evaluation results with different model architectures on CIFAR100~\cite{krizhevsky2009learning}. All values are reported in percentages, and these \emph{post hoc} methods are directly applied to the model. The best three results are highlighted with \textbf{\textcolor{red}{red}}, \textbf{\textcolor{blue}{blue}}, and \textbf{\textcolor{cyan}{cyan}}.}
    \resizebox{0.99\linewidth}{!}{
    \begin{tabular}{c|c|cc|cc|cc|cc|cc}
    \toprule
        \multirow{3}*{\textbf{Model}}&\multirow{3}*{\textbf{Methods}} & \multicolumn{2}{c|}{\textbf{iNaturalist}} & \multicolumn{2}{c|}{\textbf{SUN}} & \multicolumn{2}{c|}{\textbf{Places}} & \multicolumn{2}{c|}{\textbf{Textures}} & \multicolumn{2}{c}{\textbf{Average}}   \\
        \cmidrule{3-12}
         && FPR95 & AUROC  & FPR95 & AUROC  & FPR95 & AUROC  & FPR95 & AUROC & FPR95 & AUROC  \\
         && ($\downarrow$) & ($\uparrow$) & ($\downarrow$) & ($\uparrow$) & ($\downarrow$) & ($\uparrow$) & ($\downarrow$) & ($\uparrow$) & ($\downarrow$) & ($\uparrow$)\\
    \midrule
    \multirow{9}*{\textbf{RepVGG-A0~\cite{ding2021repvgg}}}&MSP~\cite{hendrycks2016baseline} 
    &61.55&85.03&91.05&69.19&65.45&\textbf{\textcolor{cyan}{82.10}}&86.68&65.56&76.18&75.47\\
    &ODIN~\cite{liang2017enhancing} 
    &50.20&87.88&88.00&66.56&61.85&79.34&84.87&63.89&71.23&74.42\\
    &Energy~\cite{liu2020energy} 
    &53.71&84.59&86.71&66.58&59.71&78.64&84.57&63.88&71.18&73.42\\
    &Mahalanobis~\cite{lee2018simple} 
    &81.43&74.81&89.77&67.12&79.49&73.06&64.95&82.19&78.91&74.30\\
    &GradNorm~\cite{huang2021importance} 
    &78.87&68.21&95.10&44.73&66.25&75.41&92.98&43.83&83.30&58.05\\
    &ReAct~\cite{sun2021react} 
    &48.09&\textbf{\textcolor{blue}{93.00}}&\textbf{\textcolor{cyan}{73.87}}&\textbf{\textcolor{blue}{78.12}}&\textbf{\textcolor{cyan}{61.63}}&78.43&75.23&81.36&64.71&\textbf{\textcolor{cyan}{82.73}}\\
    \cmidrule{2-12}
     \rowcolor{gray!20} \cellcolor{white}&\textbf{RankFeat}&\textbf{\textcolor{blue}{40.19}}&88.06&\textbf{\textcolor{blue}{70.47}}&76.35&\textbf{\textcolor{blue}{57.75}}&\textbf{\textcolor{blue}{83.58}}&\textbf{\textcolor{cyan}{52.89}}&\textbf{\textcolor{cyan}{83.28}}&\textbf{\textcolor{blue}{55.33}} &\textbf{\textcolor{blue}{82.82}}\\
    \cmidrule{2-12}
    \rowcolor{gray!20} \cellcolor{white} & \textbf{RankWeight} &\textbf{\textcolor{cyan}{46.16}} &\textbf{\textcolor{cyan}{89.90}}&75.18&\textbf{\textcolor{cyan}{77.63}}&77.80&75.24&\textbf{\textcolor{blue}{53.58}}&\textbf{\textcolor{blue}{85.57}}&\textbf{\textcolor{cyan}{63.18}}&82.09  \\
    \rowcolor{gray!20} \cellcolor{white} & \textbf{RankFeat} + \textbf{RankWeight} &\textbf{\textcolor{red}{19.87}} &\textbf{\textcolor{red}{95.37}} &\textbf{\textcolor{red}{56.98}} &\textbf{\textcolor{red}{81.45}} &\textbf{\textcolor{red}{41.96}} &\textbf{\textcolor{red}{88.91}} &\textbf{\textcolor{red}{35.17}} &\textbf{\textcolor{red}{91.68}} &\textbf{\textcolor{red}{38.50}} &\textbf{\textcolor{red}{89.35}} \\
    \midrule
    \multirow{9}*{\textbf{ResNet-56~\cite{he2016deep}}}&MSP~\cite{hendrycks2016baseline} 
    &77.69&78.25&93.54&66.93&81.57&76.71&88.47&65.79&85.32&71.92\\
    &ODIN~\cite{liang2017enhancing} 
    &66.92&79.25&95.05&50.45&77.45&72.88&90.51&53.47&82.48&64.01\\
    &Energy~\cite{liu2020energy} 
    &65.24&79.13&95.05&49.33&77.10&72.32&90.39&52.68&81.95&63.37\\
    &Mahalanobis~\cite{lee2018simple} 
    &89.47&69.32&91.38&54.76&82.32&77.53&68.83&79.64&83.00&70.31\\
    &GradNorm~\cite{huang2021importance} 
    &96.72&42.09&94.19&47.97&94.61&48.09&89.14&50.18&93.67&47.08\\
    &ReAct~\cite{sun2021react} 
    &50.59&\textbf{\textcolor{blue}{90.56}}&69.23&\textbf{\textcolor{blue}{85.79}}&\textbf{\textcolor{cyan}{55.38}}&\textbf{\textcolor{cyan}{87.98}}&82.60&75.51&64.50&\textbf{\textcolor{cyan}{84.96}}\\
    \cmidrule{2-12}
    \rowcolor{gray!20} \cellcolor{white} &\textbf{RankFeat} &\textbf{\textcolor{cyan}{34.62}}&88.21&\textbf{\textcolor{cyan}{61.82}}&80.50&\textbf{\textcolor{blue}{53.79}}&\textbf{\textcolor{blue}{89.71}}&\textbf{\textcolor{cyan}{30.89}}&\textbf{\textcolor{blue}{91.31}}&\textbf{\textcolor{blue}{45.28}}&\textbf{\textcolor{blue}{87.43}}\\
    \cmidrule{2-12}
    \rowcolor{gray!20} \cellcolor{white} &\textbf{RankWeight} &\textbf{\textcolor{blue}{31.93}} &\textbf{\textcolor{cyan}{88.42}} &\textbf{\textcolor{blue}{60.58}} &\textbf{\textcolor{cyan}{80.98}} &67.26 &76.96 &\textbf{\textcolor{blue}{28.51}} &\textbf{\textcolor{cyan}{90.98}} &\textbf{\textcolor{cyan}{47.07}} &84.34 \\
    \rowcolor{gray!20} \cellcolor{white} &\textbf{RankFeat} + \textbf{RankWeight} &\textbf{\textcolor{red}{15.67}} &\textbf{\textcolor{red}{93.12}} &\textbf{\textcolor{red}{53.78}} &\textbf{\textcolor{red}{88.91}} &\textbf{\textcolor{red}{39.95}} &\textbf{\textcolor{red}{90.85}} &\textbf{\textcolor{red}{22.74}} &\textbf{\textcolor{red}{92.71}} &\textbf{\textcolor{red}{33.04}} &\textbf{\textcolor{red}{91.40}} \\
    \bottomrule
    \end{tabular}
    }
    \label{tab:results_cifar}
\end{table}

%46.16 89.90|75.18 77.63|77.80 75.24|53.58 85.57| 
%43.31 88.92|86.70 66.30|83.54 67.90|83.69 64.26| 

We also evaluate our method on the CIFAR benchmark with various model architectures. The evaluation OOD datasets are the same as those of the ImageNet-1k benchmark. We take ResNet-56~\cite{he2016deep} and RepVGG-A0~\cite{ding2021repvgg} pre-trained on ImageNet-1k as the backbones, and then fine-tune them on CIAR100~\cite{krizhevsky2009learning} for $100$ epochs. The learning rate is initialized with $0.1$ and is decayed by $10$ every $30$ epoch. Notice that this training process is to obtain a well-trained classifier but the ODO methods (including ours) are still \emph{post hoc} and do not need any extra training. 

Table~\ref{tab:results_cifar} compares the performance against the \emph{post hoc} baselines. Our \texttt{RankFeat} + \texttt{RankWeight} establishes the \textit{state-of-the-art} performances across architectures on most datasets and metrics, outperforming the second best method by \textbf{26.21 \%} in the average FPR95 on RepVGG-A0 and by \textbf{31.46 \%} in the average FPR95 on ResNet-56. Applying \texttt{RankFeat} and \texttt{RankWeight} separately also achieves impressive results. Since the CIFAR images are small in resolution (\emph{i.e.,} $32{\times}32$), the downsampling times and the number of feature blocks of the original models are reduced. We thus only apply \texttt{RankFeat} to the final feature before the GAP layer.

\subsection{One-class CIFAR10}

To further demonstrate the applicability of our method, we follow~\cite{emmott2013systematic,golan2018deep,tack2020csi} and conduct experiments on one-class CIFAR10. The setup is as follows: we choose one
of the classes as the ID set while keeping other classes as OOD sets. Table~\ref{tab:oneclass_cifar10} reports the average AUROC on CIFAR10. Our \texttt{RankFeat} and \texttt{RankWeight} outperform other baselines on most subsets as well as on the average result.

\begin{table}[htbp]
    \centering
    \caption{The average AUROC (\%) on one-class CIFAR10 based on ResNet-56.}
    \resizebox{0.99\linewidth}{!}{
    \begin{tabular}{c|cccccccccc|c}
    \toprule
         Methods & Plane & Car & Bird & Cat & Deer & Dog & Frog & Horse & Ship & Truck & Mean \\
    \midrule
         MSP    &59.75 &52.48 &62.96 &48.73 &59.15 &52.39 &67.33 &59.34 &54.55 &51.97 &56.87 \\
         Energy &83.12 &91.56 &68.99 &56.02 &75.03 &77.33 &69.50 &88.41 &82.88 &84.74 &77.76 \\
         ReAct  &82.24 &96.69 &78.32 &76.84 &76.11 &86.80 &86.15 &90.95 &89.91 &94.17 &85.82 \\
         \rowcolor{gray!20}\textbf{RankFeat} &79.26 &\textbf{98.54} &82.04 &80.28 &82.89 &90.28 &89.06 &95.30 &94.11 &94.02 &88.58 \\
         \midrule
         \rowcolor{gray!20}\textbf{RankWeight} &81.32 &97.65 &84.17 &79.68 &80.54 &89.18 &90.34 &95.76 &93.78 &94.54 &88.70 \\
         \rowcolor{gray!20}\textbf{RankFeat}+\textbf{RankWeight} &\textbf{84.56} &98.15 &\textbf{85.73} &\textbf{81.43} &\textbf{84.32} &\textbf{91.78} &\textbf{90.73} &\textbf{96.21} &\textbf{95.07} &\textbf{94.78} &\textbf{90.28} \\
    \bottomrule
    \end{tabular}
    }
    \label{tab:oneclass_cifar10}
\end{table}

%\begin{table}[htbp]
%    \centering
%    \caption{\textcolor{red}{The average AUROC (\%) on one-class CIFAR100 based on ResNet-56.}}
%    \begin{tabular}{c|c}
%    \toprule
%         Methods & Mean AUROC \\
%    \midrule
%         MSP & 66.19\\
%         Energy & 73.42\\
%         ReAct & 79.58\\
%         \rowcolor{gray!20}\textbf{RankFeat}  & \textbf{82.43}\\
%    \bottomrule
%    \end{tabular}
%    \label{tab:oneclass_cifar100}
%\end{table}

%\section{Detailed Mathematical Derivation}
%\noindent \textbf{Igeood~\cite{gomes2022igeood}.} Gomes~\emph{et al.}~\cite{gomes2022igeood} proposed to use the Fisher-Rao distance~\cite{atkinson1981rao} to measure the distance between feature distributions of ID and OOD data. Let $q_{\theta}$ and $p_{\theta}$ denote the distribution of ID feature and the distribution of OOD feature, respectively. The distance is computed as:
%\begin{equation}
%    d_{\rm FR}(q_{\theta}, p_{\theta})=2\arccos{\Big(\sum\sqrt{q_{\theta}p_{\theta}}\Big)}
%\end{equation}
%where the summation is performed on the entire distribution of ID data and OOD data. Based on the distance measure, the authors further proposed to perturb the data in the input and feature space to better distinguish different distributions. 

%\section{Dataset Specifications}

%\section{Exemplary Singular Value Distribution of ID and OOD Data}

\section{Baseline Methods}

For the convenience of audiences, we briefly recap the previous \emph{post hoc} methods for OOD detection. Some implementation details of the methods are also discussed.

\noindent \textbf{MSP~\cite{hendrycks2016baseline}.} One of the earliest work considered directly using the Maximum Softmax Probability (MSP) as the scoring function for OOD detection. Let $f(\cdot)$ and $\mathbf{x}$ denote the model and input, respectively. The MSP score can be computed as:
\begin{equation}
    \texttt{MSP}(\mathbf{x}) = \max\Big({\rm Softmax}(f(\mathbf{x}))\Big)
\end{equation}
Despite the simplicity of this approach, the MSP score often fails as neural networks could assign arbitrarily high confidences to the OOD data~\cite{nguyen2015deep}.

\noindent \textbf{ODIN~\cite{liang2017enhancing}.} Based on MSP~\cite{hendrycks2016baseline}, ODIN~\cite{liang2017enhancing} further integrated temperature scaling and input perturbation to better separate the ID and OOD data. The ODIN score is calculated as:
\begin{equation}
    \texttt{ODIN}(\mathbf{x}) = \max\Big({\rm Softmax}(\frac{f(\mathbf{\bar{x}})}{T})\Big)
\end{equation}
where $T$ is the hyper-parameter temperature, and $\mathbf{\bar{x}}$ denote the perturbed input. Following the setting in~\cite{huang2021importance}, we set $T{=}1000$. According to~\cite{huang2021importance}, the input perturbation does not bring any performance improvement on the ImageNet-1k benchmark. Hence, we do not perturb the input either.

\noindent \textbf{Energy score~\cite{liu2020energy}.} Liu \emph{et al.}~\cite{liu2020energy} argued that an energy score is superior than the MSP because it is theoretically aligned with the input probability density, \emph{i.e.,} the sample with a higher energy correspond to data with a lower likelihood of occurrence. Formally, the energy score maps the logit output to a scalar function as:
\begin{equation}
    \texttt{Energy}(\mathbf{x}) = \log\sum^{C}_{i=1}\exp(f_{i}(\mathbf{x}))
\end{equation}
where $C$ denotes the number of classes.
%In line with the convention that the ID data has a higher score, the negative energy score is used.

\noindent \textbf{Mahalanobis distance~\cite{lee2018simple}.} Lee~\emph{et al.}~\cite{lee2018simple} proposed to model the Softmax outputs as the mixture of multivariate Gaussian distributions and use the Mahalanobis distance as the scoring function for OOD uncertainty estimation. The score is computed as:
\begin{equation}
    \texttt{Mahalanobis}(\mathbf{x}) = \max_{i}\Big(-(f(\mathbf{x})-\mu_{i})^{T}\Sigma(f(\mathbf{x})-\mu_{i})\Big)
\end{equation}
where $\mu_{i}$ denotes the feature vector mean, and $\Sigma$ represents the covariance matrix across classes. Following~\cite{huang2021importance}, we use $500$ samples randomly selected from ID datasets and an auxiliary tuning dataset to train the logistic regression and tune the perturbation strength $\epsilon$. For the tuning dataset, we use FGSM~\cite{goodfellow2014explaining} with a perturbation size of 0.05 to generate adversarial examples. The selected $\epsilon$ is set as $0.001$ for ImageNet-1k.

\noindent \textbf{GradNorm~\cite{huang2021importance}.} Huang~\emph{et al.}~\cite{huang2021importance} proposed to estimate the OOD uncertainty by utilizing information extracted from the gradient space. They compute the KL divergence between the {\rm Softmax} output and a uniform distribution, and back-propagate the gradient to the last layer. Then the vector norm of the gradient is used as the scoring function. Let $\mathbf{w}$ and $\mathbf{u}$ denote the weights of last layer and the uniform distribution. The score is calculated as:
\begin{equation}
    \texttt{GradNorm}(\mathbf{x}) = ||\frac{\partial D_{KL}(\mathbf{u}||{\rm Softmax}(f(\mathbf{x})))}{\partial \mathbf{w}}||_{1}
\end{equation}
where $||\cdot||_{1}$ denotes the $L_{1}$ norm, and $D_{KL}(\cdot)$ represents the KL divergence measure.

\noindent \textbf{ReAct~\cite{sun2021react}.} In~\cite{sun2021react}, the authors observed that the activations of the penultimate layer are quite different for ID and OOD data. The OOD data is biased towards triggering very high activations, while the ID data has the well-behaved mean and deviation. In light of this finding, they propose to clip the activations as:
\begin{equation}
    f_{l{-}1}(\mathbf{x})=\min(f_{l{-}1}(\mathbf{x}),\tau)
\end{equation}
where $f_{l{-}1}(\cdot)$ denotes the activations for the penultimate layer, and $\tau$ is the upper limit computed as the $90$-th percentile of activations of the ID data. Finally, the Energy score~\cite{liu2020energy} is computed for estimating the OOD uncertainty.

\noindent \textbf{ASH~\cite{djurisic2023extremely}.} Following \texttt{ReAct}~\cite{sun2021react}, \texttt{ASH} considers more diverse activation-shaping approaches beyond clipping. They consider three variants to perturb the activations:
\begin{itemize}
    \item \texttt{ASH-P}: the activations are pruned according to the threshold calculated as the $p$-th percentile of activations; 
    \item \texttt{ASH-B}: after clipping and zeroing out the activations, all non-zero values are scaled by the sparsity of activations;
    \item \texttt{ASH-S}: after clipping and zeroing out the activations, all non-zero values are scaled by a hyper-parameter calculated as the ratio between the sum of non-zero activations and the sparsity of the activations;
\end{itemize}
We observe that on the ImageNet-1k benchmark, \texttt{ASH-S} achieves the best performance. We also use \texttt{ASH-S} and compute hyper-parameters from the statistics of the ID set. 

\noindent \textbf{VRA~\cite{xu2023vra}.} Motivated by \texttt{ReAct}~\cite{sun2021react}, \texttt{VRA} leverages the variational formulation to seek the optimal activation function by extra suppression and amplification. The modified activation function takes the form:
\begin{equation}
    \texttt{VRA}(z)= \begin{cases} 0, & z <\alpha\\
    z + \gamma, & \alpha < z < \beta.\\
    \beta, & z>\beta
    \end{cases}
\end{equation}
where $z$ denotes the original activation, $\alpha$, $\beta$, and $\gamma$ are the hyper-parameters that control the thresholds and the extents of suppression/amplification. We compute these hyper-parameters based on the variational objective of \texttt{VRA}. 

Interestingly, motivated by our \texttt{RankFeat}, \texttt{VRA} also conducts some similar upper bound analyses. They show that their upper bound can be formulated as:
\begin{equation}
    \texttt{VRA}(z)\leq k_2 ||\mathbf{W}||_p||\mathbf{z}||_p + k_2 ||\mathbf{b}||_p + \log(Q)
\end{equation}
where $k_2$ is a constant upper-bounding the relation $||f(z)||_{\infty} <k_2||f(z)||_p$. According to the above inequality, maximizing $\mathbb{E}_{in}[z]-\mathbb{E}_{out}[z]$ is also equivalent to maximizing the upper bound gap between ID and OOD data. 

%\section{Impact of Removing Rank-1 Matrices on Classification Consistency}

\section{Visualization about RankFeat}

\begin{figure}[htbp]
    \centering
    \includegraphics[width=0.99\linewidth]{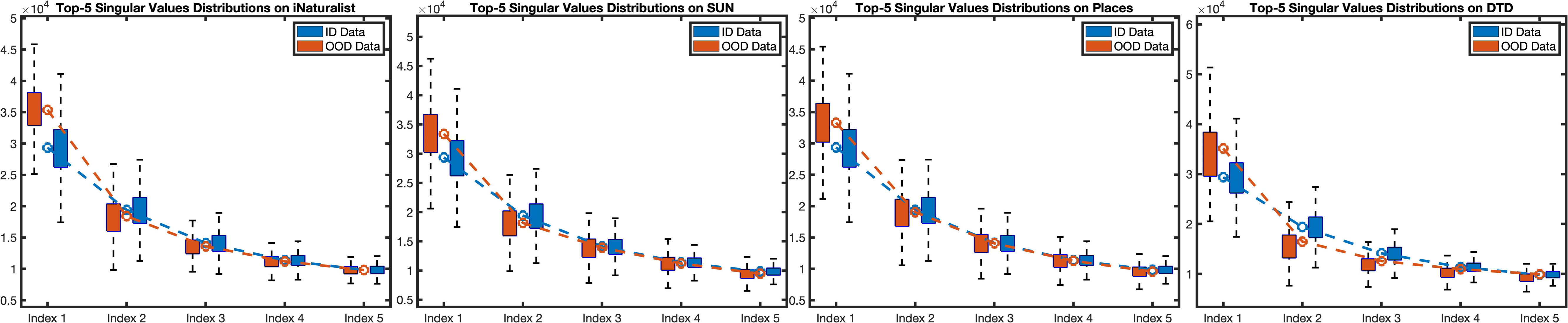}
    \caption{The top-5 singular value distribution of the ID dataset and OOD datasets. The first singular values $\mathbf{s}_{1}$ of OOD data are consistently much larger than those of ID data on each OOD dataset.}
    \label{fig:singu_dist}
\end{figure}

\subsection{Singular Value Distributions}

Fig.~\ref{fig:singu_dist} compares the top-5 singular value distribution of ID and OOD feature matrices on all the datasets. Our novel observation consistently holds for every OOD dataset: the dominant singular value $\mathbf{s}_{1}$ of OOD feature always tends to be significantly larger than that of ID feature.

\begin{figure}[htbp]
    \centering
    \includegraphics[width=0.99\linewidth]{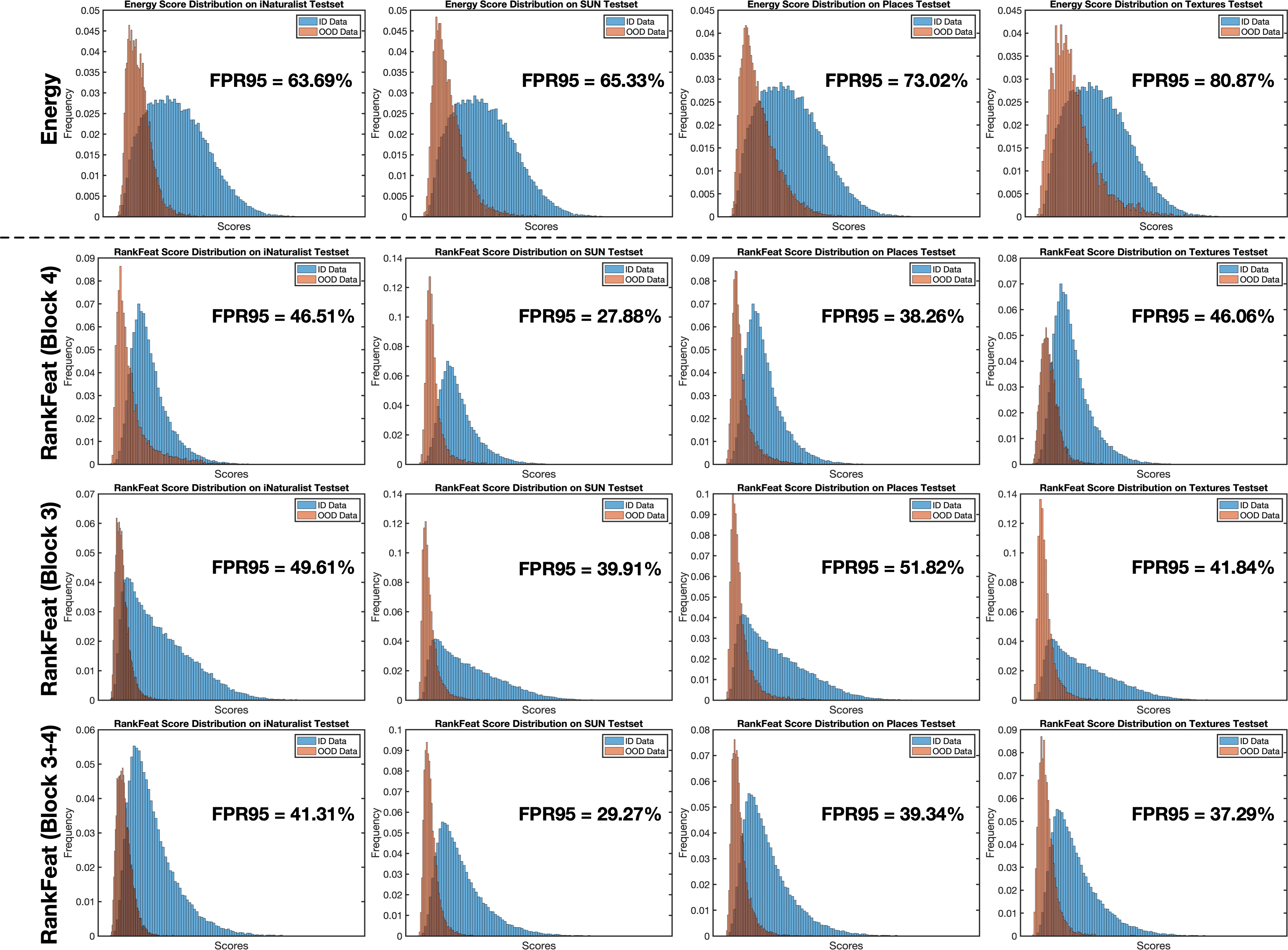}
    \caption{The score distributions of \texttt{Energy}~\cite{liu2020energy} (top row) and our proposed \texttt{RankFeat} (rest rows) on four OOD datasets. Our \texttt{RankFeat} applies to different high-level features at the later depths of the network, and their score functions can be further fused. }
    \label{fig:score_dist_complete}
\end{figure}

\subsection{Score Distributions}

Fig.~\ref{fig:score_dist_complete} displays the score distributions of \texttt{RankFeat} at Block 3 and  Block 4, as well as the fused results. Our \texttt{RankFeat} works for both high-level features. For the score fusion, when Block 3 and Block 4 features are of similar scores $(diff.{<}5\%)$, the feature combination could have further improvements.  

\subsection{Output Distributions}

Fig.~\ref{fig:output_logit}(a) presents the output distribution (\emph{i.e.,} the logits after \texttt{Softmax} layer) on \texttt{ImageNet} and \texttt{iNaturalist}. After our \texttt{RankFeat}, the OOD data have a larger reduction in the probability output; most of OOD predictions are of very small probabilities (${<}0.1$).

\subsection{Logit Distributions}

\begin{figure}[htbp]
    \centering
    \includegraphics[width=0.99\linewidth]{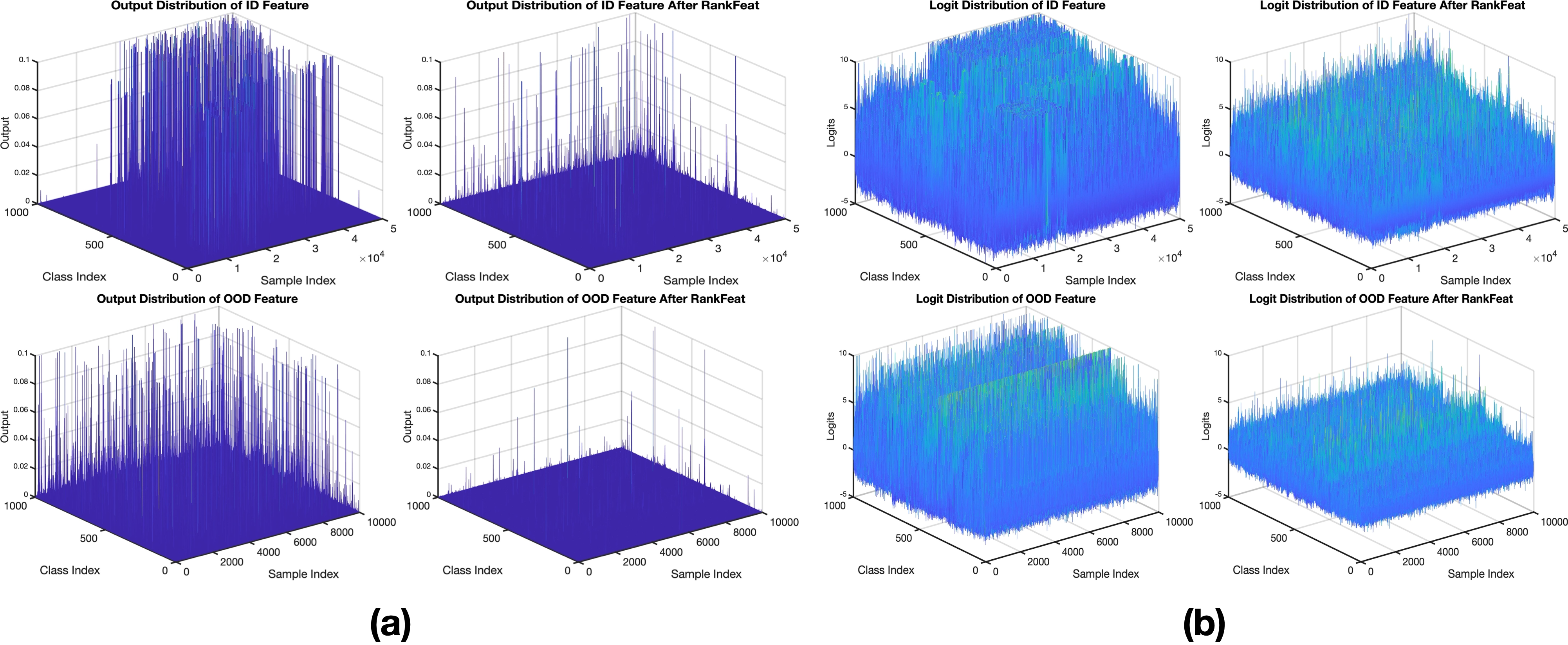}
    \caption{\textbf{(a)} Output distributions of \texttt{RankFeat}. \textbf{(b)} Logit distributions of \texttt{RankFeat}. }
    \label{fig:output_logit}
\end{figure}

Fig.~\ref{fig:output_logit}(b) displays the logits distribution of our \texttt{RankFeat}. The OOD logits after \texttt{RankFeat} have much less variations and therefore are closer to the uniform distribution.

\section{Why are the singular value distributions of ID and OOD features different?}

%It remains unclear why the singular value distributions of the ID and OOD features are different. 
In the paper, we give some theoretical analysis to explain the working mechanism of our \texttt{RankFeat}. It would be also interesting to investigate why the singular value distributions of the ID and OOD features are different. Here we give an intuitive conjecture. Since the network is well trained on the ID training set, when encountered with ID data, the feature matrix is likely to be more informative. Accordingly, more singular vectors would be active and the matrix energies spread over the corresponding singular values, leading to a more flat spectrum. On the contrary, for the unseen OOD data, the feature is prone to have a more compact representation, and less singular vectors might be active. In this case, the dominant singular value of OOD feature would be larger and would take more energies of the matrix. The informativeness can also be understood by considering applying PCA on the feature matrix. Suppose that we are using PCA to reduce the dimension of ID and OOD feature to $1$. The amount of retained information can be measured by explained variance (\%). The metric is defined as $\sum_{i=0}^{k}\mathbf{s}_{i}^2/\sum_{j=0}^{n}\mathbf{s}_{j}^2$ where $k$ denotes the projected dimension and $n$ denotes the total dimension. It measures the portion of variance that the projected data could account for. We compute the average explained variance of all datasets and present the result in Table~\ref{tab:explained_varaince}.

\begin{table}[h]
    \centering
    \caption{The average explained variance ratio (\%) of the ID and OOD datasets.}
     \resizebox{0.99\linewidth}{!}{
    \begin{tabular}{c|c|c|c|c|c}
    \toprule
         Dataset & ImageNet-1k & iNaturalist & SUN & Places & Textures  \\
    \midrule
         Explained Variance (\%) & \textbf{28.57} & 38.74 & 35.79 & 35.17 & 42.21 \\
    \bottomrule
    \end{tabular}
    }
    \label{tab:explained_varaince}
\end{table}

As can be observed, the OOD datasets have a larger explained variance ratio than the ID dataset. \emph{That being said, to retain the same amount of information, we need fewer dimensions for the projection of OOD features. This indicates that the information of OOD feature is easier to be captured and the OOD feature matrix is thus less informative.}

\textit{As for how the training leads to the difference, we doubt that the well-trained network weights might cause and amplify the gap in the dominant singular value of the ID and OOD feature.} To verify this guess, we compute the singular values distributions 
of the Google BiT-S ResNetv2-100 model~\cite{he2016identity,kolesnikov2020big} with different training steps, as well as a randomly initialized network as the baseline. 

\begin{figure}[htbp]
    \centering
    \includegraphics[width=0.99\linewidth]{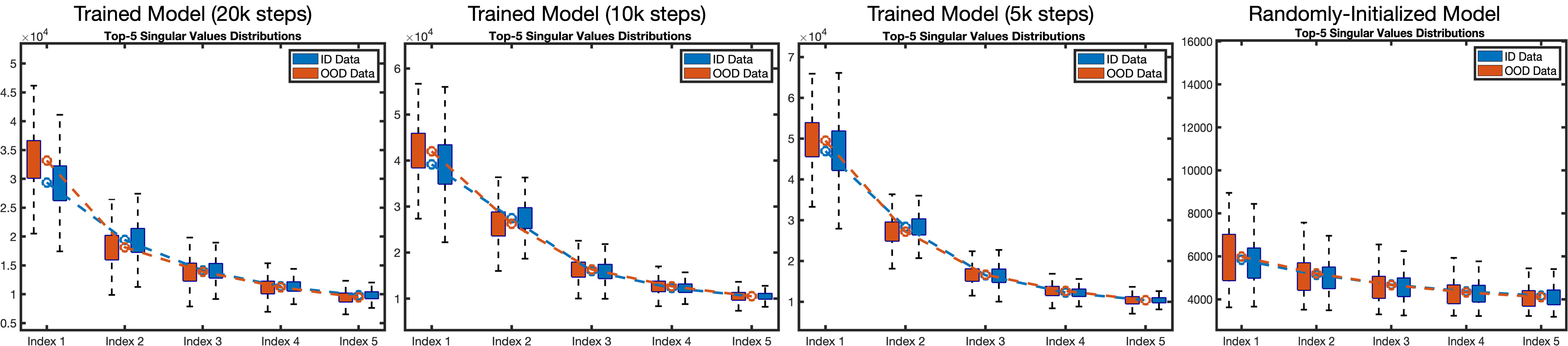}
    \caption{The top-5 largest singular value distributions of the pre-trained network with different training steps. For the untrained network initialized with random weights, the singular values distributions of ID and OOD features exhibit very similar behaviors. As the training step increases, the difference between the largest singular value is gradually amplified.}
    \label{fig:rand_train}
\end{figure}

Fig.~\ref{fig:rand_train} depicts the top-5 largest singular value distributions of the network with different training steps. Unlike the trained networks, the untrained network with random weights has quite a similar singular value distribution for the ID and OOD data. The singular values of both ID and OOD features are of similar magnitudes with the untrained network. However, when the number of training steps is increased, the gap of dominant singular value between ID and OOD feature is magnified accordingly.
This phenomenon supports our conjecture that the well-trained network weights cause and amplify the difference of the largest singular value. Interestingly, our finding is coherent with~\cite{vaze2021open}. In~\cite{vaze2021open}, the authors demonstrate that the classification accuracy of a model is highly correlated with its ability of OOD detection and open-set recognition. Training a stronger model could naturally improve the OOD detection performance. We empirically show that the gap of the dominant singular value is gradually amplifying as the training goes on, which serves as supporting evidence for~\cite{vaze2021open}. 

On the theoretical side, Huh~\emph{et al.}~\cite{huhlow2023} reveal that in general over-parameterized neural networks are biased towards finding low-rank embeddings for input data.
Further, Andriushchenko~\emph{et al.}~\cite{andriushchenko2023sharpness} present theoretical guarantees that using Sharpness-Aware Minimization (SAM)~\cite{foret2021sharpness}
provably leads to low-rank features for a diverse set of architectures. Although the conjecture has only been proved for SAM, it is known that many of its properties are generalized to other commonly used optimizers. These works provide some other theoretical evidence to support our argument.

\section{Theorem and Proof of Manchenko-Pastur Law}

In the paper, we use the MP distribution of random matrices to show that removing the rank-1 matrix makes the statistics of OOD features closer to random matrices. For self-containment and readers' convenience, here we give a brief proof of Manchenko-Pastur Law.

\begin{thm}
Let $\mathbf{X}$ be a random matrix of shape $t{\times}n$ whose entries are random variables with $E(\mathbf{X}_{ij}=0)$ and $E(\mathbf{X}_{ij}^2=1)$. Then the eigenvalues of the sample covariance $\mathbf{Y}=\frac{1}{n}\mathbf{X}\mathbf{X}^{T}$ converges to the probability density function: $\rho(\lambda) = \frac{t}{n} \frac{\sqrt{(\lambda_{+}-\lambda)(\lambda-\lambda_{-})}}{2\pi\lambda\sigma^2}\ for\  \lambda\in[\lambda_{-},\lambda_{+}]$ where $\lambda_{-}{=}\sigma^{2} (1-\sqrt{\frac{n}{t}})^2$ and $ \lambda_{+}{=}\sigma^{2} (1+\sqrt{\frac{n}{t}})^2$.
\end{thm}
\begin{proof}

Similar with the deduction of our bound analysis, the sample covariance $\mathbf{Y}$ can be written as the sum of rank-1 matrices:
\begin{equation}
    \mathbf{Y}=\sum_{s=0}^{t}=\mathbf{Y}^{s}_{n},\ \mathbf{Y}^{s}_{n}=\mathbf{U}^{s}_{n}\mathbf{D}^{s}_{n}(\mathbf{U}^{s}_{n})^{*}
\end{equation}
where $\mathbf{U}^{s}_{n}$ is a unitary matrix, and $\mathbf{D}^{s}_{n}$ is a diagonal matrix with the only eigenvalue $\beta=\nicefrac{n}{t}$ for large $n$ (rank-1 matrix). Then we can compute the Stieltjes transform of each $\mathbf{Y}^{s}_{n}$ as:
\begin{equation}
    s_{n}(z)=\frac{1}{n}{\rm tr}(\mathbf{Y}^{s}_{n}-z\mathbf{I})^{-1}
\end{equation}
Relying on Neumann series, the above equation can be re-written as:
\begin{equation}
\begin{aligned}
    s_{n}(z)&=-\frac{1}{n}\sum_{k=0}^{\infty}\frac{{\rm tr}(\mathbf{Y}^{s}_{n})^t}{z^{k+1}}\\
    &=-\frac{1}{n}\Big(\frac{n}{z}+\sum_{k=1}^{\infty}\frac{\beta^{k}}{z^{k+1}}\Big)\\
    &=-\frac{1}{n}\Big(\frac{n-1}{z}+\frac{1}{z-\beta}\Big)
\end{aligned}
\end{equation}
Let $z:=z_{n}(s)$ and we can find the function inverse of the transform:
\begin{equation}
    n s z_{n}(s)^2-n(s\beta-1)z_{n}(s)-(n-1)\beta=0
\end{equation}
The close-formed solution is calculated as:
\begin{equation}
\begin{aligned}
    z_{n}(s) &= \frac{n(s\beta-1)\pm\sqrt{n^2(s\beta-1)^2 + 4n(n-1)s\beta}}{2ns}\\
             &\approx \frac{1}{2ns}\Big(n(s\beta-1)\pm\Big|n(s\beta+1)-\cancel{\frac{2s\beta}{\beta+1}}\Big|\Big)
\end{aligned}
\end{equation}
For large $n$, the term $\frac{2s\beta}{\beta+1}$ is sufficiently small and we can omit it. The solution is defined as:
\begin{equation}
    z_{n}(s) = - \frac{1}{s} + \frac{\beta}{n(1+s\beta)}
\end{equation}
The R transform of each $\mathbf{Y}^{s}_{n}$ is given by:
\begin{equation}
    R_{\mathbf{Y}^{s}_{n}}(s) = z_{n}(-s)-\frac{1}{s}=\frac{\beta}{n(1-s\beta)} 
\end{equation}
Accordingly, the R transform for $\mathbf{Y}_{n}$ is given by:
\begin{equation}
    R_{\mathbf{Y}}(s) = t R_{\mathbf{Y}^{s}_{n}}(s) = \frac{\beta t}{n(1-s\beta)}=\frac{1}{1-s\beta}
\end{equation}
Thus, the inverse Stieltjes transform of $\mathbf{Y}$ is
\begin{equation}
    z(s)=-\frac{1}{s} + \frac{1}{1+s\beta}
\end{equation}
Then the Stieltjes transform of $\mathbf{Y}$ is computed by inverting the above equation as:
\begin{equation}
    s(z)=\frac{-(z+\beta+1)+\sqrt{(z+\beta+1)^2-4\beta z}}{2z\beta}
\end{equation}
Since $\beta=\nicefrac{b}{t}$, finding the limiting distribution of the above equation directly gives the Manchenko-Pastur distribution:
\begin{equation}
\begin{gathered}
    \rho(\lambda) = \frac{t}{n} \frac{\sqrt{(\lambda_{+}-\lambda)(\lambda-\lambda_{-})}}{2\pi\lambda\sigma^2}\ for\  \lambda\in[\lambda_{-},\lambda_{+}], \\
    \lambda_{-}{=}\sigma^{2} (1-\sqrt{\frac{n}{t}})^2, \lambda_{+}{=}\sigma^{2} (1+\sqrt{\frac{n}{t}})^2
\end{gathered}
\end{equation}
The theorem is thus proved.
\end{proof}

%5k  steps 75.75;  82.39/58.99 81.36/63.07 86.17/57.20 31.33/92.56   70.31/67.96
%10k steps 78.71;  78.45/61.80 79.33/63.47 84.18/57.23 24.91/93.43   66.72/68.98
%20k                                                                 39.69/87.84
%\subsection{Score Upper Bound}

%\subsection{Exemplary Singular Value Distribution of ID and OOD Data}